\documentclass[sigconf]{acmart}
\AtBeginDocument{%
  \providecommand\BibTeX{{%
    \normalfont B\kern-0.5em{\scshape i\kern-0.25em b}\kern-0.8em\TeX}}}
    
\usepackage{makecell}
\usepackage{multirow}
\theoremstyle{definition}

\newtheorem{definition}{Definition}[section]

\copyrightyear{2022}
\acmYear{2022}
\setcopyright{acmcopyright}
\acmConference[SIGIR '22]{Proceedings of the 45th International ACM SIGIR Conference on Research and Development in Information Retrieval}{July 11--15, 2022}{Madrid, Spain}
\acmBooktitle{Proceedings of the 45th International ACM SIGIR Conference on Research and Development in Information Retrieval (SIGIR '22), July 11--15, 2022, Madrid, Spain}
\acmPrice{15.00}
\acmDOI{10.1145/3477495.3531720}
\acmISBN{978-1-4503-8732-3/22/07}

\begin{document}
\fancyhead{}
\title{Space4HGNN: A Novel, Modularized and Reproducible Platform to Evaluate Heterogeneous Graph Neural Network}

\author{Tianyu Zhao}
\email{tyzhao@bupt.edu.cn}
\affiliation{%
  \institution{Beijing University of Posts and Telecommunications}
  \country{China}}

\author{Cheng Yang}
\email{yangcheng@bupt.edu.cn}
\affiliation{
  \institution{Beijing University of Posts and Telecommunications}
  \institution{Peng Cheng Laboratory}
  \country{China}
  }

\author{Yibo Li}
\email{liushiliushi@bupt.edu.cn}
\affiliation{%
  \institution{Beijing University of Posts and Telecommunications}
  \country{China}
}

\author{Quan Gan}
\email{quagan@amazon.com}
\affiliation{%
  \institution{AWS Shanghai AI Lab}
%   \state{Shanghai}
  \country{China}
}

\author{Zhenyi Wang}
\email{zy\_wang@bupt.edu.cn}
\affiliation{
  \institution{Beijing University of Posts and Telecommunications}
  \country{China}
}

\author{Fengqi Liang}
\email{lfq@bupt.edu.cn}
\affiliation{
  \institution{Beijing University of Posts and Telecommunications}
  \country{China}
}

\author{Huan Zhao}
\email{zhaohuan@4paradigm.com}
\affiliation{
  \institution{4Paradigm Inc.}
%   \state{Beijing}
  \country{China}
}

% \author{Hui Wang}
% \email{wangh06@pcl.ac.cn}
% \affiliation{
%   \institution{Peng Cheng Laboratory}
%   \country{China}
% }

\author{Yingxia Shao}
\email{shaoyx@bupt.edu.cn}
\affiliation{
  \institution{Beijing University of Posts and Telecommunications}
  \country{China}
}

\author{Xiao Wang}
\email{xiaowang@bupt.edu.cn}
\affiliation{%
  \institution{Beijing University of Posts and Telecommunications}
  \institution{Peng Cheng Laboratory}
  \country{China}
}

\author{Chuan Shi$^\dag$}
\thanks{$^\dag$ The corresponding author}
\email{shichuan@bupt.edu.cn}
\affiliation{%
  \institution{Beijing University of Posts and Telecommunications}
  \institution{Peng Cheng Laboratory}
  \country{China}
}

% \affiliation{%

% \affiliation{%

% \affiliation{%

% \affiliation{%

% \affiliation{%

% \affiliation{%

% \affiliation{%

\renewcommand{\shortauthors}{Zhao and Yang, et al.}

\begin{abstract}
Heterogeneous Graph Neural Network (HGNN) has been successfully employed in various tasks, but we cannot accurately know the importance of different design dimensions of HGNNs due to diverse architectures and applied scenarios. Besides, in the research community of HGNNs, implementing and evaluating various tasks still need much human effort. To mitigate these issues, we first propose a unified framework covering most HGNNs, consisting of three components: heterogeneous linear transformation, heterogeneous graph transformation, and heterogeneous message passing layer. Then we build a platform Space4HGNN by defining a design space for HGNNs based on the unified framework, which offers modularized components, reproducible implementations, and standardized evaluation for HGNNs. Finally, we conduct experiments to analyze the effect of different designs. With the insights found, we distill a condensed design space and verify its effectiveness.
\end{abstract}

\ccsdesc[500]{Computing methodologies~Machine learning}
\ccsdesc[500]{Computing methodologies~Neural networks}
\ccsdesc[500]{Information systems~Evaluation of retrieval results}

\keywords{heterogeneous graph; graph neural networks; design space}


\maketitle

\begin{figure}
    \centering
    \includegraphics[width=\columnwidth]{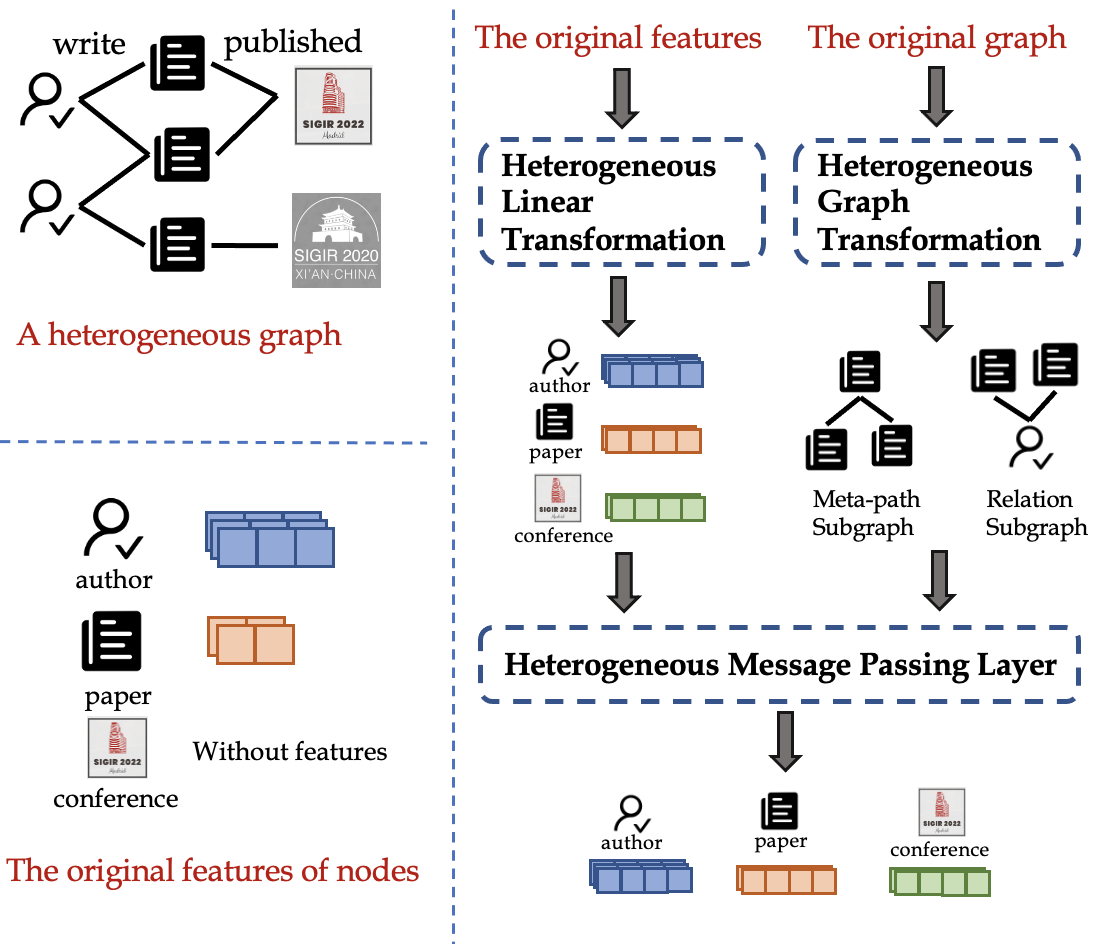}
    \caption{(left) An illustration of a heterogeneous graph and the corresponding nodes features. (right) The overall framework contains three components.}
    \label{fig:architecture}
\end{figure}
\section{Introduction}
In information retrieval, graph neural network (GNN) as graph learning and representation method has been applied in recommendation~\cite{chang2020bundle, wang2019neural, chen2021structured, chang2021sequential, zhao2017meta} and knowledge representation~\cite{gong2020attentional, cao2021dekr, hao2020dynamic, mei2018link}. Most GNNs focus on homogeneous graphs, while more and more research \cite{RGCN, HAN, jin2020learning, fang2019m, yu2021heterogeneous} shows that real world with complex interactions, e.g., social network \cite{tang2008arnetminer, wang2019online}, can be better modeled by heterogeneous graphs (a.k.a., heterogeneous information networks). Taking recommender system as an example, it can be regarded as a bipartite graph consisting of users and items, and a lot of auxiliary information also has a complex network structure, which can be naturally modeled as a heterogeneous graph. Besides, some works \cite{guo2015automatic, liu2020heterogeneous, bi2020heterogeneous, jiang2018cross, ghazimatin2020explaining, HGNNrec, HGCF, GHCF} have achieved state-of-the-art (SOTA) performance by designing heterogeneous graph neural network (HGNN). In fact, HGNNs can utilize the complex structure and rich semantic information \cite{wang2020survey}, and have been widely applied in many fields, such as e-commerce \cite{zhao2019intentgc, ji2021large}, and security \cite{sun2020hgdom, hu2019cash}.

\begin{table*}[htbp]
  \caption{Different perspectives to categorize HGNN models: (1) The first row from the perspective of neighbors to be aggregated is mentioned in Section ~\ref{pre-HGNN}. (2) The second row shows we roughly divide the existing models into three categories mentioned in Section ~\ref{aggregation paradigm}. (3) The third and the fourth rows show the components of our unified framework in Section ~\ref{framework}. (4) The fifth row: from the perspective of implementation, they contain different convolution layers. }
  \begin{center}
  \resizebox{\textwidth}{!}{
  \begin{tabular}{l|c|c|c|c}
    \toprule
        \textbf{Neighbors} & \multicolumn{3}{c|}{\textbf{One-hop}}  & \textbf{Meta-path }  \\
    % \midrule
    %     \textbf{Message} & Type-agnostic & \multicolumn{3}{c}{Type-specific} \\
    % \midrule
    %     \textbf{Reduce} & \multicolumn{2}{c|}{direct} & \multicolumn{2}{c}{Type-specific} \\
    % \midrule
    %     \textbf{Target Graph} & \multicolumn{2}{c|}{Whole graph} & \multicolumn{2}{c}{Subgraph} \\
    % \midrule
    %     \textbf{Macro} & \multicolumn{2}{c|}{None} & \multicolumn{2}{c}{Mean-Sum-Max} \\
    \midrule
        \textbf{Model Family} & \multicolumn{2}{c|}{Homogenization} & Relation & Meta-path\\
    \midrule
        \textbf{\makecell[l]{Heterogeneous Graph\\ Transformation}} & \multicolumn{2}{c|}{Homogenization of the heterogeneous graph} & Relation subgraph extraction & Meta-path subgraph extraction \\
    \midrule
        \textbf{\makecell[l]{Heterogeneous \\ Message Passing}} & \multicolumn{2}{c|}{Direct-aggregation} & \multicolumn{2}{c}{Dual-aggregation} \\
    % \midrule
    %     \textbf{\makecell[l]{Heterogeneous \\ Message Passing}} & \makecell[c]{Homogeneous \\ Direct-aggregation} & \makecell[c]{Heterogeneous \\Direct-aggregation} & \multicolumn{2}{c}{Dual-aggregation} \\
    % \midrule
    %     \multirow{2}{*}{\textbf{\makecell[l]{Heterogeneous\\ Message Passing}}} & \multicolumn{2}{c|}{Direct-aggregation} & \multicolumn{2}{c}{Dual-aggregation} \\
    \midrule
        \textbf{Graph Convolution}& \makecell[c]{Single graph \\homogeneous convolution} & \makecell[c]{Single graph \\heterogeneous convolution} & \multicolumn{2}{c}{\makecell[c]{Multiple homogeneous graph\\ convolutions applied to different subgraphs}} \\
    \midrule
        \textbf{Model} & \makecell[c]{GCN~\cite{GCN}, GAT~\cite{GAT}, \\ GraphSage~\cite{sage}, GIN~\cite{GIN}}& \makecell[c]{HGAT~\cite{HGAT}, HetSANN~\cite{HetSANN} \\HGT~\cite{HGT}, Simple-HGN~\cite{Simple-HGN}} & RGCN~\cite{RGCN}, HGConv~\cite{HGConv} & HAN~\cite{HAN}, HPN~\cite{HPN} \\
    \bottomrule
  \end{tabular}}
  \label{tab:model}
  \end{center}
\end{table*}

However, it is increasingly difficult for researchers in the field to compare existing methods and contribute with novel ones. The reason is that previous evaluations are conducted from the point of view of model-level, and we cannot accurately know the importance of each component due to diverse architecture designs and applied scenarios. 
% To figure out the effectiveness of different design architectures, we should evaluate them from the sight of module-level. 
% However, existing HGNNs' diverse implementations and complicated architectures take great challenges to define the design dimensions of HGNNs. To mitigate the challenge, 
To evaluate them from the sight of module-level, we first propose a unified framework of existing HGNNs that consists of three key components through systematically analyzing their underlying graph data transformation and aggregation procedures, as shown in Figure~\ref{fig:architecture} (right). The first component \emph{Heterogeneous Linear Transformation} is a general operation of HGNNs, which maps features to a shared feature space.
% Before performing message passing, some existing HGNN models (i.e., HAN \cite{HAN}) will explicitly transform the original graph into meta-path subgraphs, and other models (i.e., RGCN \cite{RGCN}, HGConv \cite{HGConv}) implicitly extracts relation subgraphs instead. 
Summarizing the transformed graphs used in different HGNNs, we abstract the second component \emph{Heterogeneous Graph Transformation} containing relation subgraph extraction, meta-path subgraph extraction, and homogenization of the heterogeneous graph. With that, we can explicitly decouple the selection procedure of receptive field and message passing procedure. Hence the third component \emph{Heterogeneous Message Passing Layer} can focus on the key procedure involving diverse graph convolution layers. As shown in Table~\ref{tab:model}, our framework both categorizes existing approaches and facilitates the exploration of novel ones. 
%Also, the modularized design for the framework can offer unique design dimensions that contain some high-level architectural designs or design principles of HGNNs.

With the help of the unified framework, we propose to define a design space for HGNNs, which consists of a Cartesian product of different design dimensions following GraphGym \cite{you2020design}. In GraphGym, there have been analysis results of design dimensions for GNNs, which is not enough for HGNNs. To figure out whether the guidelines distilled from GNNs are effective, our design space still contains common design dimensions with GraphGym. Besides, to capture heterogeneity, we distill three model families according to \emph{Heterogeneous Graph Transformation} in our unified framework. Based on the design space, we build a platform \textsl{Space4HGNN}~\footnote{\url{https://github.com/BUPT-GAMMA/Space4HGNN}}, which offers reproducible model implementation, standardized evaluation for diverse architecture designs, and easy-to-extend API to plug in more architecture design options. We believe \textsl{Space4HGNN} can greatly facilitate the research field of HGNNs. Specifically, we could check the effect of the tricky designs or architecture design quickly, innovate the HGNN models easily, and apply HGNN in other interesting scenarios. In addition, the platform can be used as the basis of neural architecture search for HGNNs in future work.
% ~\footnote{\url{https://github.com/BUPT-GAMMA/Space4HGNN}}

With the platform \textsl{Space4HGNN}, we conduct extensive experiments and aim to analyze the design dimensions. We first evaluate common design dimensions used in GraphGym with uniform random search, and find that they are partly effective in HGNNs. More importantly, to accurately judge the diverse architecture designs in HGNNs, we comprehensively analyze unique design dimensions in HGNNs.
And we sum up the following insights:
\begin{itemize}
\item Different model families have different suitable scenarios. The meta-path model family has an advantage in node classification task, and the relation model family performs outstandingly in link prediction task.
% \item The meta-path model family is useful with good meta-paths, and the homophily of subgraphs extracted by meta-paths may be a reference for the meta-paths selection in node classification task.
\item The preference for different design dimensions may be opposite in different tasks. For example, node classification task prefers to apply L2 Normalization and remove Batch Normalization. However, the better choices of the same datasets for link prediction task are the opposite.
\item We should select graph convolution carefully which varies greatly across datasets. Besides, the design dimensions like the number of message passing layers, hidden dimension and dropout are all important.
\end{itemize}

Finally, we distill a condensed design space according to the analysis results, whose scale is reduced by 500 times. We evaluate it in a new benchmark HGB \cite{Simple-HGN} and demonstrate the effectiveness of the condensed design space.

And we sum up the following contributions:
\begin{itemize}
    \item As far as we know, we are the first to propose a unified framework and define a design space for HGNNs. They offer us a module-level sight and help us evaluate the influences of different design dimensions, such as high-level architectural designs, and design principles.
    \item We release a platform \textsl{Space4HGNN} for design space in HGNNs, which offers modularized components, standardized evaluation, and reproducible implementation of HGNN. We conduct extensive experimental evaluations to analyze HGNNs comprehensively, and provide findings behind the results based on \textsl{Space4HGNN}. 
    It allows researchers to find more interesting findings and explore more robust and generalized models.
    \item Following the findings, we distill a condensed design space. Experimental results on a new benchmark HGB \cite{Simple-HGN} show that we can easily achieve state-of-the-art performance with a simple random search in the condensed space.
\end{itemize}

\section{Related Work}
Notations and the corresponding descriptions used in the rest of the paper are given in Table~\ref{tab:notation}. More preliminaries can be found in Appendix~\ref{app:preliminary}.

\begin{table}[htpb]
    \centering
    \caption{Notation and corresponding description.}
    \begin{tabular}{c|c}
    \toprule
        Notation & Description \\
        \midrule
        $v_{i}$ & The node $v_{i}$ \\ 
        $e_{ij}$ & The edge from node $v_{i}$ to node $v_{j}$ \\
        $\mathcal{N}_{i}$ & The neighbors of node $v_{i}$ \\
        $\mathbf{h}$ & The hidden representation of a node  \\
        $\mathbf{W}$ & The trainable weight matrix \\
        $f_{v}$ & The node type mapping function \\
        $f_{e}$ & The edge type mapping function \\
        $\phi $ & The message function \\
    \bottomrule
    \end{tabular}
    \label{tab:notation}
\end{table}
\subsection{Heterogeneous Graph Neural Network}\label{pre-HGNN}
Different from GNNs, HGNNs need to handle the heterogeneity of structure and capture rich semantics of heterogeneous graphs. According to the strategies of handling heterogeneity, HGNN can be roughly classified into two categories: HGNN based on one-hop neighbor aggregation (similar to traditional GNN) and HGNN based on meta-path neighbor aggregation (to mine semantic information), shown in Table~\ref{tab:model}.

\subsubsection{\textbf{HGNN based on one-hop neighbor aggregation}} 
To deal with heterogeneity, this kind of HGNN usually contains type-specific convolution. Similar to GNNs, the aggregation procedure occurs in one-hop neighbors. As earliest work and an extension of GCN \cite{GCN}, RGCN \cite{RGCN} assigns different weight matrices to different relation types and aggregates one-hop neighbors. With many GNN variants appearing, homogeneous GNNs inspire more HGNNs and then HGConv \cite{HGConv} dual-aggregate one-hop neighbors based on GATConv \cite{GAT}. A recent work SimpleHGN \cite{Simple-HGN} designs relation-type weight matrices and embeddings to characterize the heterogeneous attention over each edge. Besides, some earlier models, like HGAT \cite{HGAT}, HetSANN \cite{HetSANN}, HGT\cite{HGT}, modify GAT \cite{GAT} with heterogeneity by assigning heterogeneous attention for either nodes or edges. 
% \quan{Here we only talk about "one-hop aggregation", so I don't think you need to talk about direct- and dual-aggregation here; you can say that in Section 4.3.}

\begin{figure*}[htpb]
    \centering
    \includegraphics[width=\textwidth]{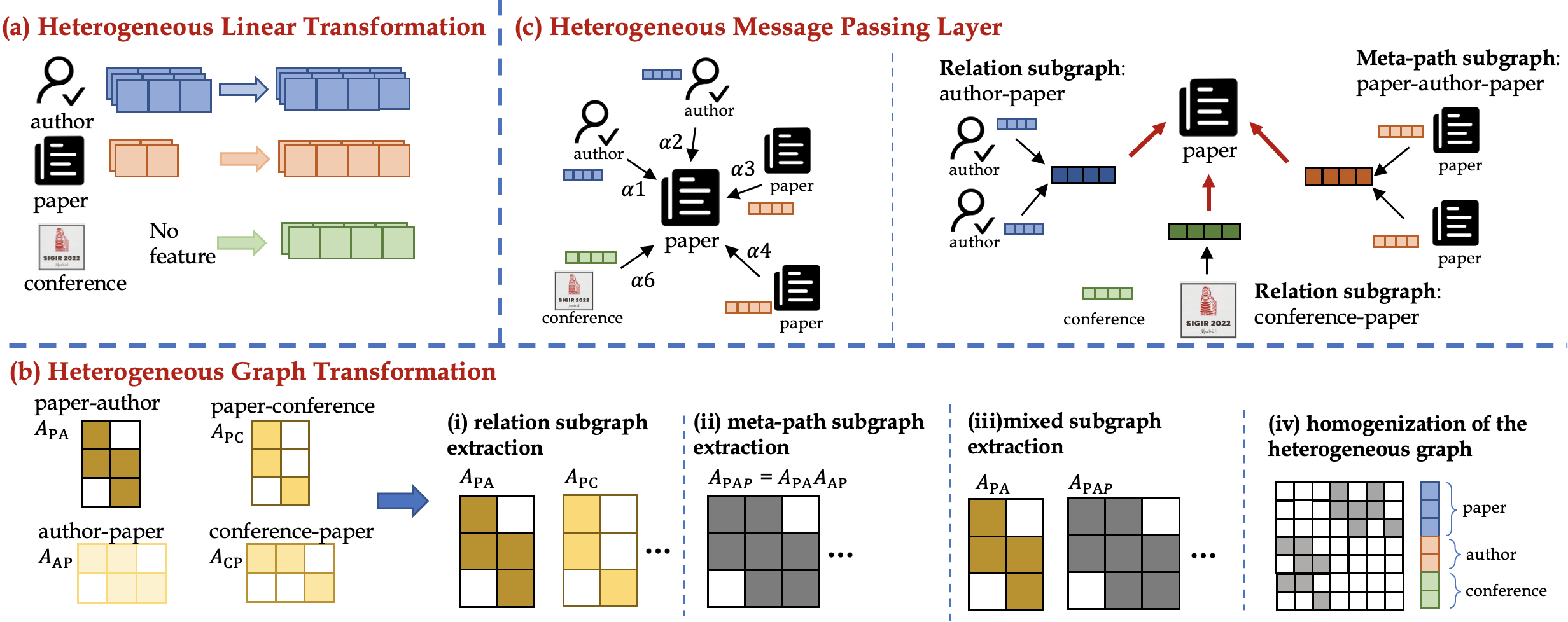}
    \caption{
    The detailed description of three components using the graph illustrated in Figure~\ref{fig:architecture}.
    (a) \emph{Heterogeneous Linear Transformation} maps all node features to a shared feature space.
    % (b) Feature preprocessor: the features of different nodes with no feature or different dimension features will be mapped to a shared feature space. 
    (b) \emph{Heterogeneous Graph Transformation} (The left one) The original graph consists of four adjacency matrices representing four relations. (The right four) Four transformation methods.
    (c) Two aggregation methods in \emph{Heterogeneous Message Passing Layer}. (The left one) The direct-aggregation: lines assigned normalized attention coefficients indicate aggregation procedure. (The right one) The dual-aggregation: the black solid line indicates the micro-level aggregation procedure applied in different subgraphs and the thick red solid lines indicates the macro-level aggregation procedure.}
    % (d) Two heterogeneous aggregation methods. (i) The direct-aggregation:  dotted lines assigned normalized attention coefficients indicates aggregation procedure. (ii) The dual-aggregation: the black solid line indicates the micro-level aggregation procedure and the thick red solid lines indicates the macro-level aggregation procedure.}
    % \quan{A few suggestions.  (1) In the figures you can just say $A_{PA}$ instead of "relation subgraph: PA" which (i) saves more space, and (ii) can more easily align to each matrix.  Also, for the first occurrence of caption "Meta-path subgraph: PAP" I would rather say "$A_{PAP} = A_{PA} A_{AP}$" instead, which explicitly says how the meta-path subgraphs are constructed.  The second "Meta-path subgraph: PAP" can be replaced with $A_{PAP}$ since you already said before how to compute it.  (2) I don't think putting feature preprocessor in the figure is necessary, as it is not directly related to our new design space dimensions, especially since Table 1 and Table 4 never talked about it.  (3) I would rather have two aggregation methods separated in the figure (with one taking the place of feature preprocessor).}
    \label{fig:graph}
\end{figure*}

\subsubsection{\textbf{HGNN based on meta-path neighbor aggregation}}Another class of HGNNs is to capture higher-order semantic information with hand-crafted meta-paths. Different from the previous, aggregation procedure occurs in neighbors connected by meta-path. As a pioneering work, HAN \cite{HAN} first uses node-level attention to aggregate nodes connected by the same meta-path and utilizes semantic-level attention to fuse information from different meta-paths. Because the meta-path subgraph ignores all the intermediate nodes, MAGNN \cite{MAGNN} aggregates all nodes in meta-path instances to ensure that information will not be missed. Though meta-paths contain rich semantic information, the selection of meta-paths needs human prior and determines the performance of HGNNs. Some works like GTN \cite{GTN} learn meta-paths automatically to construct a new graph. HAN \cite{HAN} and HPN \cite{HPN}, which are easy to extend, will be included in our framework for generality. 
%   \quan{Here we only talk about "meta-path aggregation", so this two-stage-thing-a-majig can also be moved to Section 4.3.  Instead, you should emphasize the aspect that HAN constructs a meta-path subgraph.  Afterwards you can say "because meta-path subgraph ignores all the intermediate nodes, MAGNN aggregates..."}

\subsection{Model Evaluation and Design Space}
There are many works to measure progress in the field by evaluating models. The work \cite{dacrema2019we} reports the results of a systematic analysis of neural recommendation mentioning HGNNs and sheds light on potential problems. Several works \cite{errica2019fair, dwivedi2020benchmarking, shchur2018pitfalls} discuss how to make fair a comparison between GNN models. In HGNNs, a recent work \cite{Simple-HGN} revisits HGNNs and proposes issues with existing HGNNs. The above work is to evaluate models from the model-level sight. 

Though rigorous theoretical understanding of neural network is not enough, it is imperative to perform empirical studies of neural network to discover better architectures. In visual recognition, some works \cite{radosavovic2019network, radosavovic2020designing}design network design spaces to help advance the understanding of network design and discover design principles that generalize across settings. Inspired by that, GraphGym \cite{you2020design} proposes a GNN design space and a GNN task space to evaluate model-task combinations comprehensively. In recommendation system, the work \cite{zhenyi} profiles the design space for GNNs based on collaborative filtering.

Here we aim to extensively explore the design space of HGNNs involving many design dimensions and evaluate different design architectures. Different from the model-level sight of \cite{Simple-HGN}, we evaluate the design dimensions from module-level sight and distill helpful design principles. Different from GraphGym \cite{you2020design}, our design space focuses on the unique design dimension of HGNNs and explores the differences with GNNs. 

\section{A Unified Framework of Heterogeneous Graph Neural Network} \label{framework}
As shown in Table~\ref{tab:model}, we categorize many mainstream HGNN models, which could be applied in many scenarios, e.g, link prediction~\cite{hao2020dynamic, mei2018link} and recommendation~\cite{bi2020heterogeneous, jiang2018cross, ghazimatin2020explaining}.
Through analyzing the underlying graph data and the aggregation procedure of existing HGNNs, we propose a unified framework of HGNN that consists of three main components: 
\begin{itemize}
\item \emph{Heterogeneous Linear Transformation} maps features or representations with heterogeneity to a shared feature space.
\item \emph{Heterogeneous Graph Transformation} offers four transformation methods for heterogeneous graph data to select the receptive field.
\item \emph{Heterogeneous Message Passing Layer} defines two aggregation methods suitable for most HGNNs.
\end{itemize}

\subsection{Heterogeneous Linear Transformation}
Due to the heterogeneity of nodes, different types of nodes have different semantic features even different dimensions. Therefore, for each type of nodes (e.g., node $v_{i}$ with node type $f_{v}\left(v_{i}\right)$), we design a type-specific linear transformation to project the features (or representations) of different types of nodes to a shared feature space. The linear transformation is shown as follows:
\begin{equation}
\mathbf{h}_{v_{i}}^{\prime}=\mathbf{W}_{f_{v}\left(v_{i}\right)} \cdot \mathbf{h}_{v_{i}},
\end{equation}
where $\mathbf{h}_{i}$ and $ \mathbf{h}_{i}^{\prime}$ are the original and projected feature of node, respectively.
As shown in Figure~\ref{fig:graph} (a), we transform node features with a type-specific linear transformation for nodes with features. Nodes without features or full of noise could be assigned embeddings as trainable vectors, which is equivalent to assigning them with a one-hot vector combined with a linear transformation. 
% As shown in Figure ~\ref{fig:graph}(b), the node type \textbf{author} and \textbf{paper} have different dimension and they will be mapped into the same space. And for node type \textbf{conference} without feature, the nodes will be assigned embeddings.

% \subsubsection{\textbf{Representation Post-processor}}
% The final representation can serve various downstream tasks, e.g., node classification and link prediction. So we need the representation post-processor to transform the representation after message passing layers. We only need to reduce the target node dimensions to the number of classes to classify in node classification task. In link prediction task, we may need to post-process two node types involving loss function with type-specific linear.

\subsection{Heterogeneous Graph Transformation}\label{HG transformation}

% \emph{Heterogeneous Graph Transformation} to offer graph data transformation (e.g. relation subgraph extraction, meta-path subgraph extraction, and homogenization of the heterogeneous graph). With that, we can explicitly decouple the selection procedure of receptive field and message passing procedure. Specifically, the gap between aggregation based on one-hop neighbor and that based on meta-path neighbor is broken. 
In previous work, aggregation based on one-hop neighbor usually applies the graph convolution layer in the original graph, which implicitly selects the one-hop (relation) receptive field. And aggregation based on meta-path neighbor is usually done on constructed meta-path subgraphs, which explicitly selects the multi-hop (meta-path) receptive field. Relation subgraphs are special meta-path subgraphs (note that the original graph is a special case of relation subgraphs). To unify both, we propose a component to abstract the selection procedure of the receptive field, which determines which nodes are aggregated. Besides, the component decouples the selection procedure of receptive field and message passing procedure introduced in the following subsection.

As shown in Figure~\ref{fig:graph} (b), we therefore designate a separate stage called \emph{Heterogeneous Graph Transformation} for graph construction, and categorize it into (i) relation subgraph extraction that extracts the adjacency matrices of the specified relations, (ii) meta-path subgraph extraction that constructs the adjacency matrices based on the pre-defined meta-paths, (iii) mixed subgraph extraction that builds both kinds of subgraphs, (iv) homogenization of the heterogeneous graph (but still preserving $f_v$ and $f_e$ for node and edge type mapping). For relation or meta-path extraction, we could construct subgraphs by specifying relation types or pre-defined meta-paths. 
% Heterogeneous Graph Transformation decouples the selection procedure of receptive field and message passing procedure introduced in the following subsection.

% \quan{Briefly explain what is relation subgraph extraction and meta-path subgraph extraction.}
%Besides, a heterogeneous graph also can be transformed into a homogeneous graph presented by a whole matrix combined a vector used to distinguish node types.
% \quan{The whole section can be condensed to something like "In most HGNN models, aggregation is usually done on either constructed meta-path subgraphs or extracted relation subgraphs (note that the original graph a special case of relation subgraphs).  We therefore designate a separate stage called \emph{Heterogeneous Graph Transformation} for graph construction, and categorize it into relation subgraph extraction, meta-path subgraph extraction, or mixed subgraph extraction that builds both kinds of graphs."}

\subsection{Heterogeneous Message Passing Layer}\label{Hmp-aggregation}
%In Section ~\ref{pre-HGNN}, we introduce a conventional way to classify HGNNs. However, this classification did not find enough commonality from the perspective of implementation, resulting in difficulties in designing space and expanding new models. In order to better design space for HGNN, we abstract two main forms of aggregation, which are direct-aggregation and dual-aggregation.
In Section ~\ref{pre-HGNN}, we introduce a conventional way to classify HGNNs. However, this classification did not find enough commonality from the implementation perspective, resulting in difficulties in defining design space and searching for new models. Therefore, we instead propose to categorize models by their aggregation methods.

\begin{table}[!htbp]
    \caption{Direct-aggregation with attention mechanism.}
    \begin{center}
    \resizebox{\columnwidth}{!}{
    \begin{tabular}{c|c}
    \toprule
        Mechanisms & Attention Coefficients $e_{i j}$ \\
    \midrule
        GAT \cite{GAT} & $\text { LeakyReLU }\left(\boldsymbol{a}^{T}\left[\mathbf{W} \mathbf{h}_{i}||\mathbf{W} \mathbf{h}_{j}\right]\right)$ \\
  \midrule
        HGAT \cite{HGAT} & $\text { LeakyReLU }\left(\boldsymbol{a}^{T} \boldsymbol{\alpha}_{T(j)} \left[\mathbf{h}_{i}|| \mathbf{h}_{j} \right]\right)$ \\
  \midrule
        HetSANN \cite{HetSANN} & $\text { LeakyReLU }\left(\boldsymbol{a}^{T}\left[\mathbf{W}_{T(i)T(i)} \mathbf{h}_{i}||\mathbf{W}_{T(i)T(j)} \mathbf{h}_{j} \right]\right)$\\
    \midrule
        HGT \cite{HGT} & $\mathbf{W}_{Q_T(i)} \mathbf{h}_{i} \mathbf{W_{\phi(e)}^{A T T}} (\mathbf{W}_{K_T(j)} \mathbf{h}_{j})^{T} $ \\
    \midrule
        Simple-HGN \cite{Simple-HGN} & $\text { LeakyReLU }\left(\boldsymbol{a}^{T}\left[\mathbf{W} \mathbf{h}_{i}\left\|\mathbf{W} \mathbf{h}_{j}\right\| \mathbf{W}_{r} \boldsymbol{r}_{\psi(\langle i, j\rangle)}\right]\right)$\\
    \bottomrule
    \end{tabular}}
    \label{tab:attention}
    \end{center}
\end{table}

\subsubsection{\textbf{Direct-aggregation}}\label{direct-aggregation}
The aggregation procedure is to reduce neighbors directly without distinguishing node types.
The basic baselines of HGNN models are GCN, GAT, and other GNNs used in the homogeneous graph. A recent work\cite{Simple-HGN} shows that the simple homogeneous GNNs, e.g., GCN and GAT, are largely underestimated due to improper settings. 

As shown in Figure~\ref{fig:graph} (c: the left one), we will explain it under the message passing GNNs formulation and take GAT \cite{GAT} as an example. The message function is $\phi=\alpha_{i j} \mathbf{h_{j}}^{(L)}, j \in \mathcal{N}_{i}$. The feature of node $i$ in $(L+1)$-th layer is defined as
\begin{equation}
\mathbf{h}_{i}^{L+1}=\sigma\left(\sum_{j \in \mathcal{N}_{i}} \alpha_{i j} \mathbf{W} \mathbf{h}_{j}^{L}\right),
\end{equation}
where $\mathbf{W}$ is a trainable weight matrix, $\mathcal{N}_{i}$ is neighbors of node $v_{i}$ and $\alpha_{i j}$ is the normalized attention coefficients between node $v_{i}$ and $v_{j}$, defined by that:

\begin{equation}\label{softmax}
\alpha_{i j}=\operatorname{softmax}_{i}\left(e_{i j}\right)=\frac{\exp \left(e_{i j}\right)}{\sum_{k \in \mathcal{N}_{i}} \exp \left(e_{i k}\right)}.
\end{equation} 
% \quan{This does not cover GCN (at least in the first glance) where $\alpha$ are not summed up to one.}

The correlation of node $v_i$ with its neighbor $j \in N_{i}$ is represented by attention coefficients $e_{i j}$. Changing the form of $e_{i j}$ yields other heterogeneous variants of GAT, which we summarize in Table~\ref{tab:attention}.

% Some typical HGNN models can be categorized into the direct-aggregation as shown in Figure ~\ref{fig:graph}. They are all heterogeneous extensions of GAT, which calculate attention coefficients considering node or edge heterogeneity. The last four of the Table ~\ref{tab:attention} are heterogeneous attention mechanisms.
% So the homogeneous convolution can be extended to a heterogeneous version by considering heterogeneity and here we define them as heterogeneous direct-aggregation. \quan{Not necessary.}

% \begin{figure}[h]
%     \centering
%     \includegraphics[width=0.5\textwidth]{images/aggregationv2.png}
%     \caption{Two heterogeneous aggregation methods. (a) The direct-aggregation:  dotted lines assigned normalized attention coefficients indicates aggregation procedure. (b) The dual-aggregation: the black solid line indicates the micro-level aggregation procedure and the thick red solid lines indicates the macro-level aggregation procedure.}
%     \label{fig:two aggregation methods}
% \end{figure}

\begin{figure*}[htpb]
    \centering
    \includegraphics[width=\textwidth]{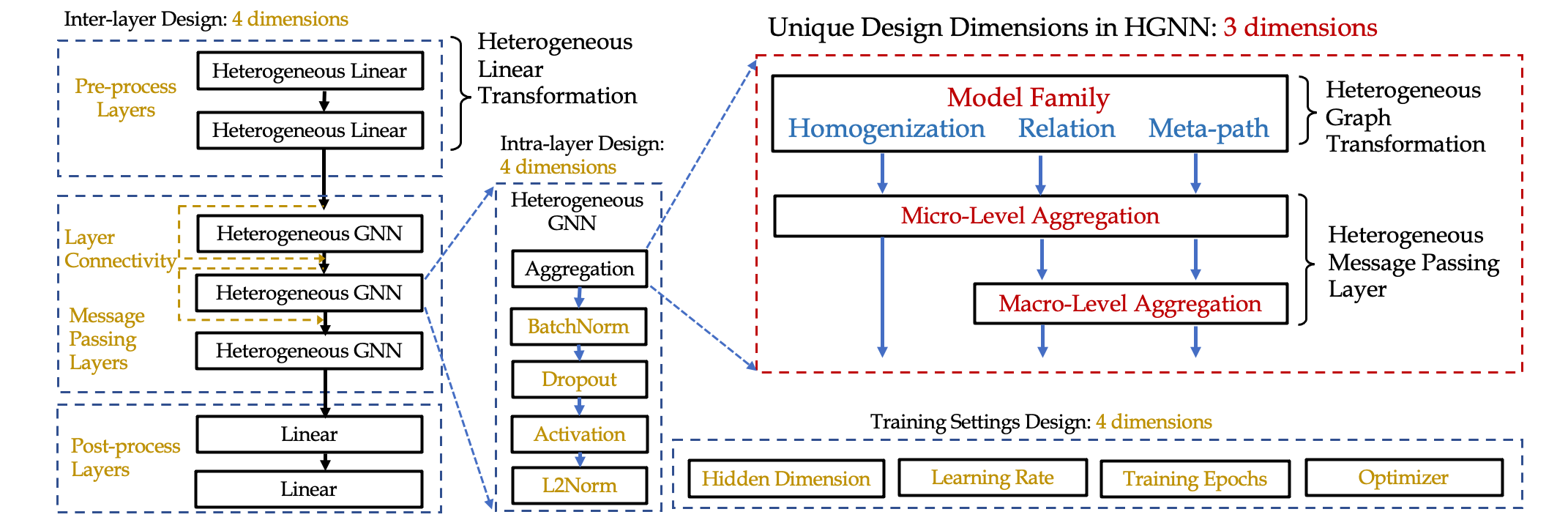}
    \caption{Design space: (1) The red and yellow font represents the design dimension and the blue font means choice of model families. (2) The red dotted frame includes the dimensions of the unique design in HGNNs. And blue dotted frame indicates some common dimensions with GraphGym.}
    \label{fig:design}
\end{figure*}

\subsubsection{\textbf{Dual-aggregation}}\label{dual-aggregation}
Following \cite{HGConv}, we define two parts of dual-aggregation: micro-level (intra-type) and macro-level (inter-type) aggregation. As shown in Figure~\ref{fig:graph} (c: the right one), micro-level aggregation is to reduce node features within the same relation, which generate  type-specific features in relation/meta-path subgraphs, and macro-level aggregation is to reduce type-specific features across different relations. When multiple relations have the same destination node types, their type-specific features are aggregated by the macro-level aggregation.  
% \quan{There is a new term "macro features" which is not used elsewhere.  Could you rephrase this sentence so you don't have to invent a new term?} 

Generally, each relation/meta-path subgraph utilizes the same micro-level aggregation (e.g., graph convolution layer from GCN or GAT). In fact, we can apply different homogeneous graph convolutions for different subgraphs in our framework. The multiple homogeneous graph convolutions combined with macro-level aggregation is another form of heterogeneous graph convolution compared with heterogeneous graph convolution in direct-aggregation. There is a minor difference between the heterogeneous graph convolution of direct-aggregation and that of dual-aggregation. 
We modify Eq.~\ref{softmax} and define it in Eq.~\ref{softmax2}, where $\mathcal{N}_{i}^{f_{v}(j)}$ means the neighbors type of node $v_i$ is the same as type of node $j$.
\begin{equation}\label{softmax2}
\alpha_{i j}=\operatorname{softmax}_{i}\left(e_{i j}\right)=\frac{\exp \left(e_{i j}\right)}{\sum_{k \in \mathcal{N}_{i}^{f_{v}(j)}} \exp \left(e_{i k}\right)}.
\end{equation}

\textbf{Example: HAN \cite{HAN} and HGConv \cite{HGConv}.} 
In HAN, the node-level attention is equivalent to a micro-level aggregation with GATConv, and the semantic-level attention is macro-level aggregation with attention, which is the same with HGConv. The HGConv uses the relation subgraphs, which means aggregating the one-hop neighbors, but HAN extracts multiple meta-path subgraphs, which means aggregating multi-hop neighbors. According to \emph{Heterogeneous Graph Transformation} in Section ~\ref{HG transformation}, the graph constructed can be a mixture of meta-path subgraphs and relation subgraphs. So the dual-aggregation can also be operated in a mixture custom of subgraphs to aggregate different hop neighbors.

% \begin{table*}
%   \caption{HGNN models category}
%   \begin{tabular}{lc|c|c}
%     \toprule
%         - & \textbf{Aggregation} & \textbf{Direct-aggregation} & \textbf{Dual-aggregation} \\
%     \midrule
%         \multirow{6}{*}{\textbf{General HG}} & gcnconv & GCN & - \\
%         & gatconv & GAT & - \\
%         & dual-attention & HGAT & - \\
%         & type-attention & HetSANN & - \\
%         & mutual-attention & HGT & - \\
%         & relation-attention & Simple-HGN & - \\
%     \midrule
%         \multirow{2}{*}{\textbf{relation extraction}} & gcnconv+sum & - & RGCN \\
%         & gatconv+attention & - & HGConv \\
%     \midrule
%         \multirow{2}{*}{\textbf{meta-path extraction}} & gatconv+attention & - & HAN \\
%         & appnpconv+attention & - & HPN \\
%     \bottomrule
%   \end{tabular}
% \label{tab:model}
% \end{table*}

\section{Design Space for Heterogeneous Graph Neural Network}
Inspired by GraphGym~\cite{you2020design}, we propose a design space for HGNN, which is built as a platform \textsl{Space4HGNN} offering modularized HGNN implementation for researchers introduced at last.   
% \quan{So our design space will indeed not cover HGT and Simple-HGN?  That's OK.  However, "well-designed" sounds positive, thus will make people wonder why you don't cover it.  I guess "unique" or "uniquely handcrafted" because you want to emphasize all those attentions are different?}.
% \quan{Consider move the last sentence up before "Our space will not contain ..." as the omission is not what we define for design space.}

\subsection{Designs in HGNN}
As illustrated in Figure~\ref{fig:design}, we will describe it from two aspects: common designs with GraphGym and unique designs distilled from HGNNs.
%Though GraphGym \cite{you2020design} distills guidelines for designing well-performing GNNs, we also design the related dimensions to figure out whether they are still effective in HGNNs. Besides, according to the unified framework proposed above, we propose three HGNN model families, which can reproduce most original HGNN models and extend new variants by applying different choices as unique dimensions in HGNN.

\begin{table}[h]
    \caption{Common design dimensions with GraphGym.}
    \begin{center}
    % \resizebox{\columnwidth}{!}{
    \begin{tabular}{c|c}
    \toprule
       \textbf{Design Dimension} & \textbf{Choices} \\
    % \midrule
    %     Paradigm & Homogenization, Relation, Meta-path \\
    %     \makecell[c]{Convolution \\ Micro-level Aggregation} & GCNConv, GATConv, SageConv, GINConv\\
    %     Macro-level Aggregation & Mean, Max, Sum, Attention \\
    \midrule
        Batch Normalization & True, False \\
        Dropout & 0, 0.3, 0.6 \\
        Activation & Relu, LeakyRelu, Elu, Tanh, PRelu \\
        L2 Normalization & True, False \\
    \midrule
        Layer Connectivity & STACK, SKIP-SUM, SKIP-CAT \\
        Pre-process Layers & 1, 2, 3 \\
        Message Passing Layers & 1, 2, 3, 4, 5, 6\\
        Post-process Layers & 1, 2, 3 \\
    \midrule
        Optimizer & Adam, SGD \\
        Learning Rate & 0.1, 0.01, 0.001, 0.0001 \\
        Training Epochs & 100, 200, 400 \\
        Hidden dimension & 8, 16, 32, 64, 128 \\
    \bottomrule
    \end{tabular}
    \label{tab:dimension}
    \end{center}
\end{table}
\subsubsection{Common Designs with GraphGym}
The common designs with GraphGym involves 12 design dimensions, categorized three aspects, intra-layer, inter-layer and training settings. The dimensions with corresponding choices are shown in Table~\ref{tab:dimension}. More detailed description is provided in Appendix~\ref{app:design}

\subsubsection{Unique Design in HGNNs} \label{aggregation paradigm}
With the unified framework, we try to transform the modular components into unique design dimensions in HGNNs. According to \cite{radosavovic2019network}, a collection of related neural network architectures, typically sharing some high-level architectural structures or design principles (e.g., residual connections), could be abstracted into a model family. With that, we distill three model families in HGNNs. 

% The \textbf{homogenization model family} uses the direct-aggregation combined with any graph convolutions. Here we use the term \emph{homogenization} because all HGNNs included here apply direct-aggregation after the homogenization of the heterogeneous graph mentioned in Section~\ref{HG transformation}. The \textbf{relation model family} applies relation subgraph extraction and dual-aggregation. The \textbf{meta-path model family} applies meta-path subgraph extraction and dual-aggregation. As shown in Table~\ref{tab:model families}, three model families involve three design dimensions with candidate choices. 

% \begin{table}[h]
%     \caption{Three model families with corresponding sub-space}
%     \begin{center}
%     \resizebox{\columnwidth}{!}{
%     \begin{tabular}{c|c|c}
%     \toprule
%       \textbf{Model Family} & \textbf{\makecell[c]{Convolution\\Micro-level Aggregation}} & \textbf{Macro-level Aggregation} \\
%     \midrule
%         Homogenization & \multirow{3}{*}{\makecell[c]{GCNConv, GATConv\\ SageConv, GINConv}}  &  -\\
%     \midrule
%         Relation & & \multirow{2}{*}{\makecell[c]{Mean, Max\\ Sum, Attention }} \\
%     \midrule
%         Meta-path \\
%         % \makecell[c]{Convolution \\ Micro-level Aggregation} & GCNConv, GATConv, SageConv, GINConv\\
%         % Macro-level Aggregation & Mean, Max, Sum, Attention \\
%     \bottomrule
%     \end{tabular}}
%     \label{tab:model families}
%     \end{center}
% \end{table}
\begin{table}[h]
    \caption{Unique design dimensions in HGNNs.}
    \begin{center}
    \resizebox{\columnwidth}{!}{
    \begin{tabular}{c|c}
    \toprule
       \textbf{Design Dimension} & \textbf{Choices} \\
    \midrule
        Model Family & Homogenization, Relation, Meta-path \\
    \midrule
        \makecell[c]{Micro-level Aggregation \\ (Graph Convolution Layer)} & GCNConv, GATConv, SageConv, GINConv\\
    \midrule
        Macro-level Aggregation & Mean, Max, Sum, Attention \\
    \bottomrule
    \end{tabular}}
    \label{tab:model families}
    \end{center}
\end{table}

\paragraph{\textbf{The Homogenization Model Family}}
The homogenization model family uses the direct-aggregation combined with any graph convolutions. Here we use the term \emph{homogenization} because all HGNNs included here apply direct-aggregation after the homogenization of the heterogeneous graph mentioned in Section~\ref{HG transformation}. Homogeneous GNNs and heterogeneous variants of GAT mentioned in Section~\ref{Hmp-aggregation} all fall into this model family. The homogeneous GNNs are usually evaluated as basic baselines in HGNN papers. Though it losses type information, it is confirmed that the simple homogeneous GNNs can outperform some existing HGNNs \cite{Simple-HGN}, which means they are nonnegligible and supposed to be seen as a model family. We select four typical graph convolution layers, which are GraphConv \cite{GCN}, GATConv \cite{GAT}, SageConv-mean \cite{sage} and GINConv\cite{GIN} as analyzed candidates.
% The heterogeneous variants of GAT are included here but they are not choices in convolution dimension since it cannot be used in the other two paradigms.  
% \quan{Do you mean our experiments did not cover all those variants?  If we extend the convolution modules to include those attentions then they will be covered, right?  If so, we can phrase it that way, like "The heterogeneous variants of GAT also fall into this category with appropriate convolution modules."}

\paragraph{\textbf{The Relation Model Family}}
The model family applies relation subgraph extraction and dual-aggregation. The first HGNN model RGCN \cite{RGCN} is a typical example in relation model family, whose dual-aggregation consists of a micro-level aggregation with SageConv-mean and macro-level aggregation of Sum.  HGConv \cite{HGConv} is a combination of GATConv and attention. We could get other designs by enumerating the combinations of micro-level and macro-level aggregation.  In our experiments, we set the micro-level aggregations the same as graph convolutions in the homogenization model family, and macro-level aggregations are chosen among Mean, Max, Sum, and Attention.
%So we could get a number of designs through flexible combination of mirco-level and macro-level aggregation. Here the micro-level aggregation is consistent with convolution in homogeneous paradigm. The candidate macro-level aggregations are mean, max, sum and attention.
% \quan{Noticed that you frequently have "And" as the first word of a sentence.  This is not necessary.}

\paragraph{\textbf{The Meta-path Model Family}}
The model family applies meta-path subgraph extraction and dual-aggregation. The instance HAN \cite{HAN} has the same dual-aggregation with HGConv \cite{HGConv} in the relation model family but different subgraph extraction. The candidate of micro-level and macro-level aggregations is the same as those in the relation model family.
% The HGNNs based on meta-path utilize the semantic information well and show great performance. However, the meta-path selection needs human prior and \cite{Simple-HGN} concludes that meta-paths are not necessary in most heterogeneous datasets. 
% \quan{We don't have to talk about how authors and other people judge metapath-based GNNs, meaning that the previous two sentences are unnecessary.  You can just say (how) HAN falls into this paradigm.} 
% Here we hope to report comprehensive analysis whether the HGNNs based on meta-paths with various dual-aggregation combination make progress.  \quan{This sentence is also not necessary.} 

\subsection{\textsl{Space4HGNN}: Platform for Design Space in HGNN} \label{space4hgnn}
We developed \textsl{Space4HGNN}, a novel platform for exploring HGNN designs. We believe Space4HGNN can significantly facilitate the research field of HGNNs. It is implemented with PyTorch~\footnote{\url{https://pytorch.org/}} and DGL~\footnote{\url{https://github.com/dmlc/dgl}}, using the OpenHGNN~\footnote{\url{https://github.com/BUPT-GAMMA/OpenHGNN}} package. It also offers a standardized evaluation pipeline for HGNNs, much like \cite{you2020design} for homogeneous GNNs. For faster experiments, we offer parallel launching. Its highlights are summarized below.

\subsubsection{\textbf{Modularized HGNN Implementation}}
The implementation closely follows the GNN design space GraphGym. It is easily extendable, allowing future developers to plug in more choices of design dimensions (e.g., a new graph convolution layer or a new macro-aggregation). Additionally, it is easy to import new design dimensions to \textsl{Space4HGNN}, such as score function in link prediction.

\subsubsection{\textbf{Standardized HGNN Evaluation}}
\textsl{Space4HGNN} offers a standardized evaluation pipeline for diverse architecture designs and HGNN models. Benefiting from OpenHGNN, we can evaluate diverse datasets in different tasks easily and offer visual comparison results presented in Section~\ref{experiments}.
% % % \input{Input/Evaluation}

\begin{table}[htpb]
    \caption{\textbf{Statistics of HGB datasets.} The prefix \emph{HGBn} and \emph{HGBl} present node classification task, link prediction task.}
    \label{tab:dataset Statistics}
    \resizebox{\columnwidth}{!}{
    \centering
    \begin{tabular}{c|c|c|c|c|c|c}
    \toprule
        Dataset & \#Nodes & \makecell[c]{\#Node \\Types} & \#Edges & \makecell[c]{\#Edge\\ Types} & \makecell[c]{Name for Node \\ Classification Task}& \makecell[c]{Name for Link\\ Prediction Task}\\
    \midrule
        DBLP & 26,128 & 4 & 239,566 & 6 & HGBn-DBLP & HGBl-DBLP  \\ 
        IMDB & 21,420 & 4 & 86,642 & 6 & HGBn-IMDB & HGBl-IMDB \\
        ACM & 10,942 & 4 & 547,872 & 8 & HGBn-ACM &  HGBl-ACM \\
        Freebase & 180,098 & 8 & 1,057,688 & 36 & HGBn-Freebase &-\\
        PubMed & 63,109 & 4 & 244,986 & 10 & HGBn-PubMed & HGBl-PubMed \\
        Amazon & 10,099 & 1 & 148,659 & 2 &-&HGBl-amazon\\
        LastFM & 20,612 & 3 & 141,521 & 3 &-&HGBl-LastFM \\

    \bottomrule
    \end{tabular}}
\end{table}

\begin{figure*}[bpth]
    \centering
    \includegraphics[width=\textwidth]{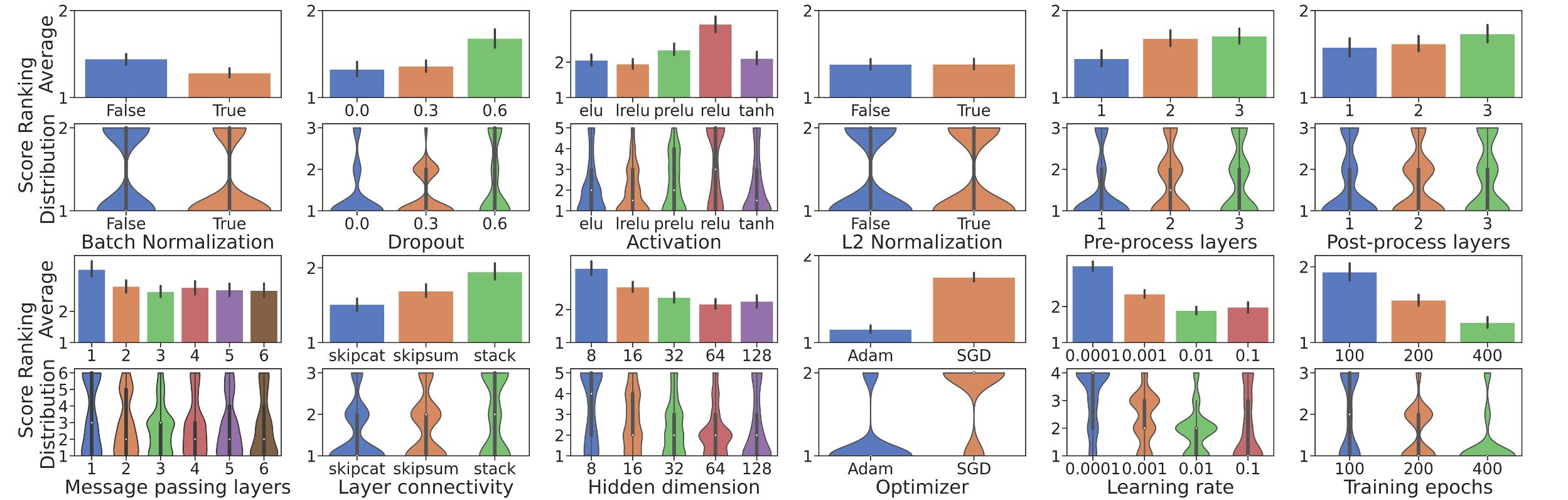}
    \caption{\textbf{Ranking analysis for design choices in 12 common design dimensions with GraphGym.} Lower is better.  %Note: (1) the macro function is only analyzed in relation paradigm and meta-path paradigm; (2) The number of heads is only analyzed in GATConv.
    }
    \label{fig:rank}
\end{figure*}

\section{Experiments}\label{experiments}
\subsection{Datasets}
% \textbf{Issues with existing datasets.}
% First, different works \cite{HAN, GTN, MAGNN} use different dataset versions. Even though they are extracted from the same original dataset, different papers make seemingly minor changes such as using different dataset filtering and data split, which could non-trivially impact model performance. 

% different papers make minor changes for them, including different remove of nodes and edges, and different data split, where the model could benefit from the minor changes.
% \quan{What is the point of this sentence?  Are we saying which dataset we are using?}
% To dealt with that, we use the Heterogeneous Graph Benchmark(HGB) \cite{Simple-HGN} as datasets for node classification and link prediction task. Besides node classification and link prediction, we process two datasets for recommendations, Yelp and DoubanMovie. For details, refer to Appedndix ~\ref{NC_LP_dataset}. 

% Second, the small data perturbations can greatly change the predictions of GNNs \cite{zugner2018adversarial}. The experiments of \cite{shchur2018pitfalls} shows that a single and fixed split of train/val/test are fragile and misleading results. Most models may run experiments many times using different random seeds for reproducibility, but the random split is usually not included.  \quan{Isn't this talking about the same issue as the first point?}

% For now, there are some benchmarks \cite{Simple-HGN, HNE, ogb} for us to evaluate the algorithms. To analyze some issues from \cite{Simple-HGN}, 

We select the Heterogeneous Graph Benchmark (HGB) \cite{Simple-HGN}, a benchmark with multiple datasets of various heterogeneity (i.e., the number of nodes and edge types). 
% The HGB is organized as a public competition, so it does not release the test label to prevent data leakage. 
To save time and submission resources, we report the test performance of the configuration with best validation performance in Table~\ref{tab:NC_performance}. Other experiments are evaluated on a validation set with three random 80-20 training-validation splits. 
The statistics of HGB are shown in Table~\ref{tab:dataset Statistics}. 
%We select five datasets (DBLP, IMDB, ACM, Freebase, PubMed) for node classification task, six datasets (DBLP, IMDB, ACM, amazon, LastFM, PubMed) for link prediction task.
% \quan{Did I get this correctly?}
%If we get performance by submitting results, our experiments will cost a amount of time and submission resources even though the HGB Team gives us a test channel to implement the evaluation of test data. So we only evaluate the best performance in the test data in Table ~\ref{tab:NC_performance}, and other experiments will spilt the original train data into train/valid 80\%/ 20\%  and repeat 3 times randomly. For dataset details, please refer to Appendix ~\ref{app:dataset}.

\subsection{Evaluation Technique} \label{random_search}
% Our design space could be viewed as a hyper-parameter space. Compared with neural networks configured by a pure grid search, random search over the same domain can find models that are as good or better within a small fraction of the computation time \cite{bergstra2012random}. Besides,
Our design space covers over 40M combinations, and a full grid search will cost too much.  We adapt controlled random search from GraphGym \cite{you2020design} setting the number of random experiments to 264, except that we ensure that every combination of dataset, model family, and micro-aggregation receives 2 hits.  We draw bar plots and violin plots of rankings of each design choice following the same practice as GraphGym. As shown in Figure~\ref{fig:rank}, in each subplot, rankings of each design choice are aggregated over all 264 setups via bar plot and violin plot. The bar plot shows the average ranking across all the 264 setups (lower is better). The violin plot indicates the smoothed distribution of the ranking of each design choice over all the 264 setups.

\subsection{Evaluation of Design Dimensions Common with GraphGym}
\subsubsection{Overall Evaluation}
\label{exp:common_design}
The evaluation results of design dimensions common with GraphGym~\cite{you2020design} are shown in Figure~\ref{fig:rank}, from which we draw the following conclusions. % The results mostly align with previous findings on GraphGym, with domain-specific differences.  %however, there exist some interesting domain-specific results. We enumerate the key experimental findings as below from two aspects, findings aligned with GraphGym and findings specific to HGNN.

\emph{\textbf{Findings aligned with GraphGym:}}
% \paragraph{\textbf{Findings aligned with GraphGym}}
\begin{itemize}
    \item We also confirmed that \textbf{BN} \cite{BN} yields better results, while \textbf{L2-Norm} did not have a significant impact. However, task-wise evaluation of both dimensions in Section~\ref{task_eva} reveals a different and more insightful story.
    % \item Over-fitting and over-smoothing are non-negligible challenges in GNNs. To mitigate the problems, the corresponding solutions are proposed. We also confirmed that \textbf{Batch Normalization} \cite{BN} facilitates neural network training and avoids over-fitting, but only in link prediction task. More results are analyzed in Section~\ref{task_eva}. Comparing with earlier HGNNs, recent models ~\cite{Simple-HGN} apply a skip connection to gain improvement. Here our evaluation confirms that \textbf{skip connection} outperforms than STACK, and SKIP-CAT is favorable. 
    \item There is no definitive conclusion for the best \textbf{number of message passing layers}; each dataset has its own best number, the same as what GraphGym observed. % The number of layers means the neighbors hops aggregated and there is few differences in average ranking between different number of message passing layers. In different dataset, each choice has had best performance.
    \item The characteristic of \textbf{training settings} (e.g., optimizer and training epochs) is similar to GraphGym.
\end{itemize}

% (1) To mitigate the problems, the corresponding solutions are proposed. As mentioned by \cite{you2020design}, we also confirmed that \textbf{Batch Normalization} \cite{BN} facilitates neural network training and avoids over-fitting. \textbf{Dropout} as a simple way to prevent neural network from over-fitting is used here to randomly drop node representations as a mean of regularization. One view \cite{you2020design} is that GNNs are thus robust to noise and outliers by neighborhood aggregation. Comparing with earlier HGNNs, recent models ~\cite{Simple-HGN} apply a skip connection to gain improvement. Here our evaluation confirms that \textbf{skip connection} outperforms than stack, and SKIP-CAT is favorable. 

% \paragraph{\textbf{Findings specific to HGNN}}
%\noindent
\emph{\textbf{Findings different from GraphGym:}}
\begin{itemize}
    \item A \textbf{single linear transformation} (pre-process layer) is usually enough. We think that this is because our heterogeneous linear transformation is node type-specific which has enough parameters to transform representations. 
    \item The widely used \textbf{activation} Relu may no longer be as suitable in HGNNs. Tanh, LeakyReLU, and ELU are better alternatives. PReLU, stood out in GraphGym, is not the best choice in our design space.
    % \item Higher hidden dimension gets better performance. %in a range of candidate choices.
    % However, higher hidden dimension means higher cost of time and space in memory. We recommend choosing as high a dimension as possible, as budget allowed. 
    \item %Dropout as a simple way to prevent the neural network from over-fitting is used here to randomly drop node representations. We think that in some datasets, applying dropout can avoid the emergence of extreme representations and makes HGNN more robust.
    Different from GraphGym, we found that \textbf{Dropout} is necessary to get better performance.  We think the reason is that parameters specific to node types and relation types lead to over-parametrization.
\end{itemize}
% (1) A single linear transformation (pre-process layer and post-process layer) is usually enough. We think that this is because our heterogeneous linear transformation is node type-specific which has enough parameters to transform representations. 
% (2) The widely used activation Relu may no longer be as suitable in HGNNs. Tanh, LeakyReLU, as well as ELU, are better alternatives. PReLU standing out in \cite{you2020design} is not the best choice in our design space.
% (3) Higher hidden dimension gets better performance in a range of candidate choices. However, higher hidden dimension means higher cost of time and space in memory. We recommend choosing as high a dimension as possible, as budget allowed. 
% (4) Dropout as a simple way to prevent the neural network from over-fitting is used here to randomly drop node representations. We think that in some datasets, applying dropout can avoid the emergence of extreme representations and makes HGNN more robust.
\begin{figure}[htpb]
    \centering
    \includegraphics[width=\columnwidth]{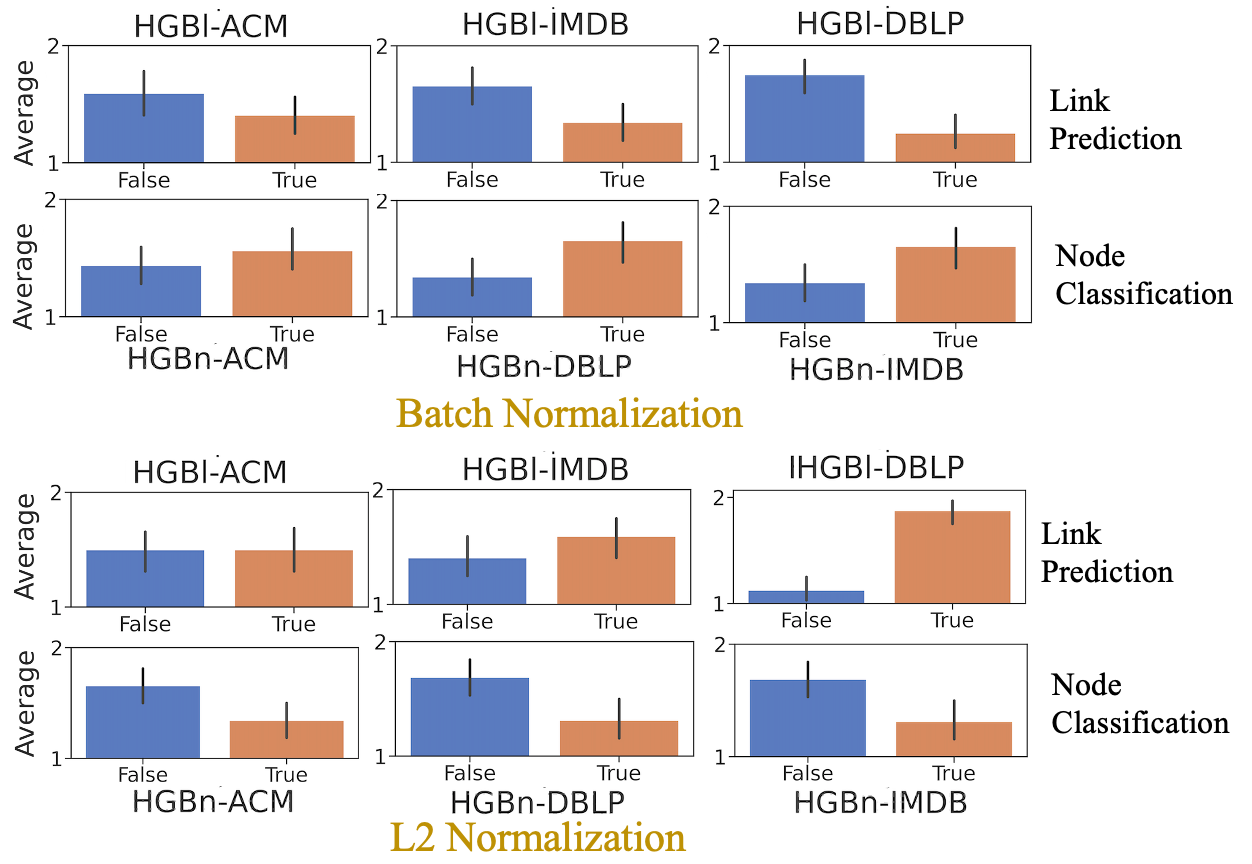}
    \caption{\textbf{Ranking analysis for design choices in BN and L2-Norm over different tasks.} Each column represents the same dataset for different tasks.}
    \label{fig:task_wise}
\end{figure}

\subsubsection{Task-wise Evaluation}\label{task_eva}
We previously observed that BN yields better performance in general. However, task-wise evaluation in Figure~\ref{fig:task_wise} showed that \textbf{BN is better on link prediction but worse on node classification}. Meanwhile, although L2-Norm does not seem to help in overall performance, it actually \textbf{performs better on node classification but worse on link prediction}.  We think that BN scales and shifts nodes according to the global information, which may lead to more similar representations and damage the performance of the node classification task, and L2-Norm scales the representation and thus the link score to [-1,1], which may invalidate the Sigmoid of the score function.

% Apart from evaluating the overall performance of each dimension aggregated from all datasets, we also evaluate from a finer granularity, that how well-performing HGNN designs differ across different tasks. As shown in Fig.\ref{fig:task_wise}, the preferable choices of Batch Normalization and L2 Normalization vary across different tasks. Specifically, datasets for node classification prefer to apply L2 Normalization (L2-Norm) and remove Batch Normalization (BN). However, the preference is the opposite in datasets for the link prediction task. BN here will scale and shift nodes according to the global information and L2Norm generally scale from node-wise without the global information. We think that the global information may lead to more similar representations, which damages the performance of the node classification task. L2Norm scales the representation and makes the link score be [-1,1] which may invalidate the Sigmoid of the score function.
 
% In some cases, the aggregation of nodes face the problem of numerical explosion which can be alleviated with L2-Norm.

\begin{figure}[htpb]
    \centering
    \includegraphics[width=\columnwidth]{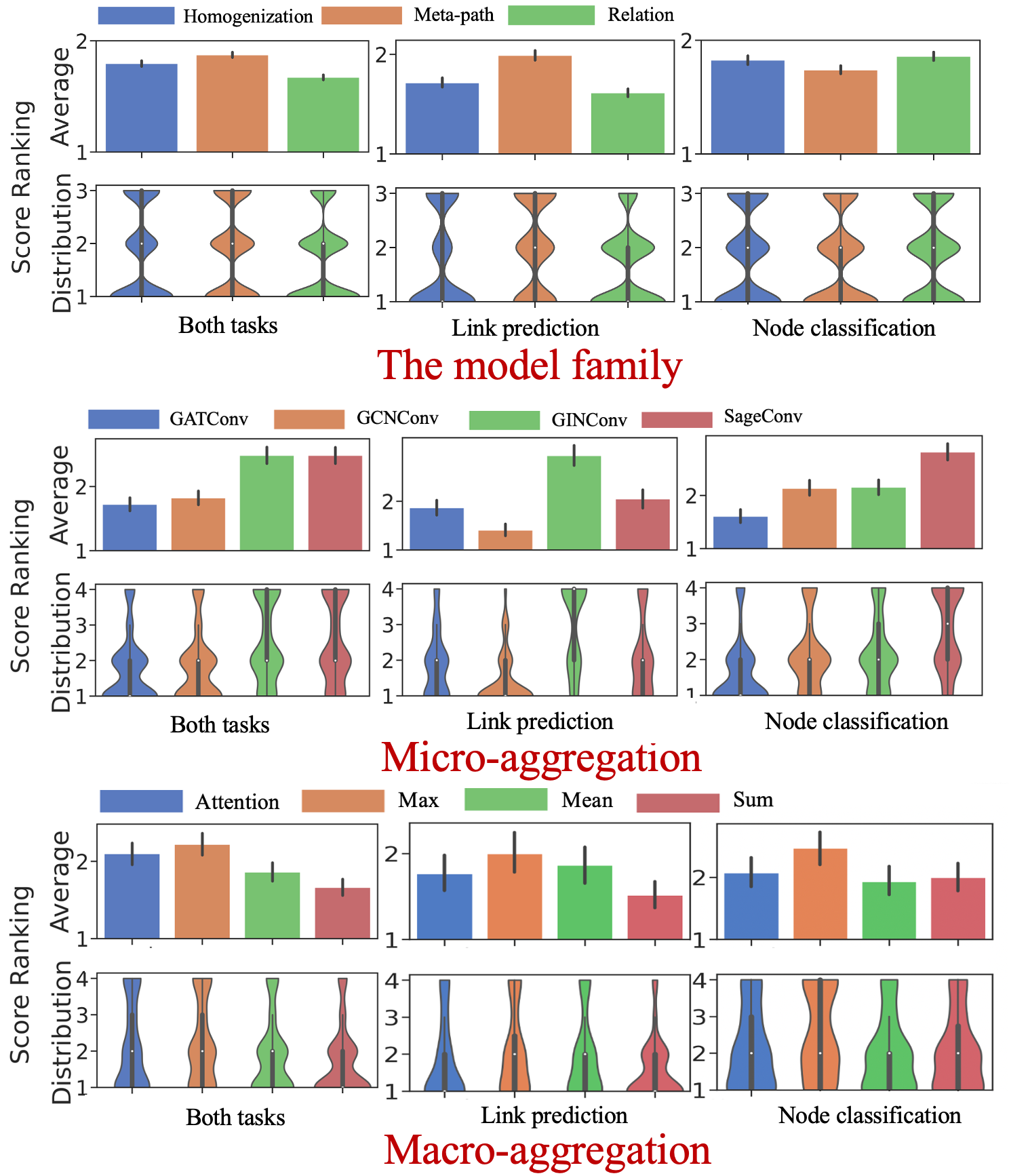}
    \caption{\textbf{Ranking analysis for design choices in 3 unique design dimensions over different tasks.}}
    \label{fig:task_rank}
\end{figure}

\begin{figure*}[htpb]
    \centering
    \includegraphics[width=\textwidth]{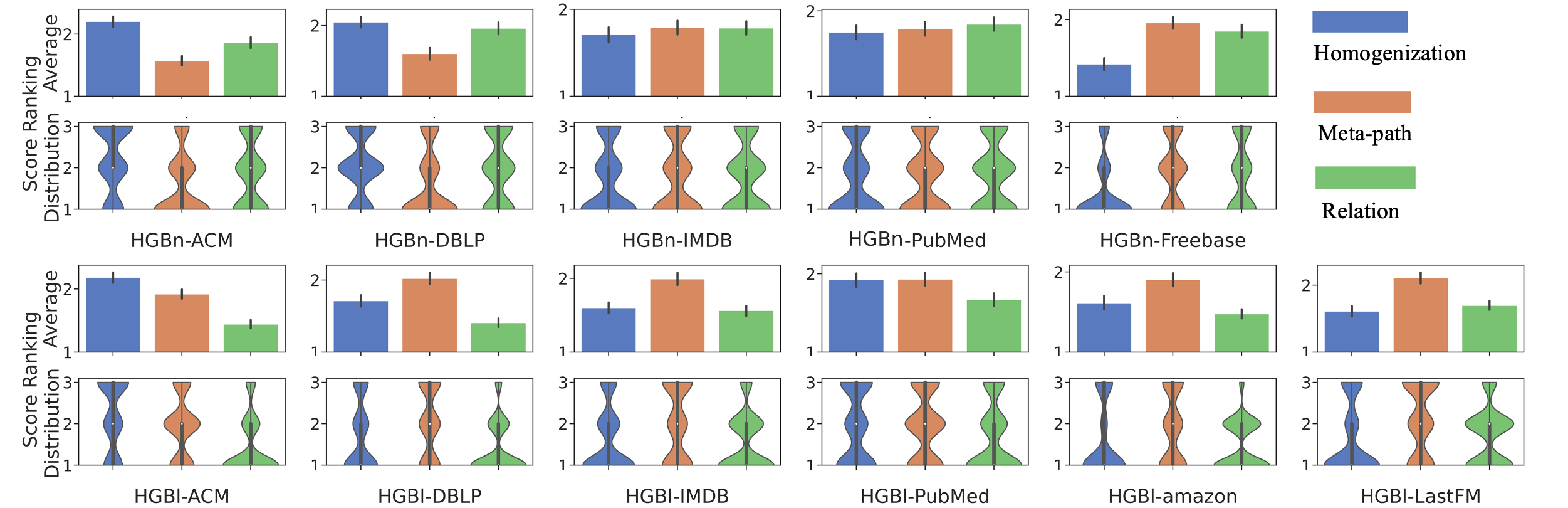}
    \caption{\textbf{Ranking analysis for different model families on different datasets.} The top is dataset of node classification task and the bottom is datasets of link prediction task. Lower is better.}
    \label{fig:rank_dataset}
\end{figure*}

% \begin{figure*}[htpb]
%     \centering
%     \includegraphics[width=\textwidth]{images/distributionv3.png}
%     \caption{\textbf{Distribution estimates for different model families in different datasets.} The intersection of the curve and the upper edge of the box represent the best performance.}
%     \label{fig:dis_dataset}
% \end{figure*}

\subsection{Evaluation of Unique Design Dimensions in HGNNs}
How to design and apply HGNN is our core issue. This section analyzes unique design dimensions in HGNNs to describe the characteristics of high-level architecture designs. From the average ranking shown in Figure~\ref{fig:task_rank}, we can see that the meta-path model family has a small advantage in the node classification task. The relation model family outperforms in aggregated results in all datasets, and the homogenization model family is competitive. For the micro-aggregation design dimension, GCNConv and GATConv are preferable for link prediction task and node classification task, respectively. For the macro-aggregation design dimension, Mean and Sum have a more significant advantage.

% According to the controlled random search mentioned in Section~\ref{random_search}, the model family is uniformly sampled and the experiments are conducted in common dimensions can be analyzed here. 

% Besides, we use score EDF, which is introduced in Section~\ref{dis_est}, to evaluate the performance distribution of the three model families on all datasets. 
% The distribution estimate results are shown in Fig.~\ref{fig:dis_dataset} and Fig.~\ref{fig:gnn_type_dis}. The discrimination between different model families is smaller than that between different micro-aggregation design dimensions. 
% \begin{figure*}[htpb]
%     \centering
%     \includegraphics[width=\textwidth]{images/rank_gnn_type.png}
%     \caption{\textbf{Ranking analysis for the micro-aggregation design dimension on different datasets.} Different micro-aggregation vary greatly across datsest.}
%     \label{fig:gnn_type_dis}
% \end{figure*}

\subsubsection{\textbf{The Model Family}}

% From the average ranking shown in Fig.~\ref{fig:task_rank}, we can see that the meta-path model family has a small advantage in the node classification task. The relation model family outperforms in aggregated results in all datasets and the homogenization model family is competitive.
% We aim to analyze them from the sight of module-level to determine the effects of different design dimensions. 
%The rank analysis in Fig.~\ref{fig:task_rank} shows performance on the whole in different tasks.
To more comprehensively describe the corresponding characteristics of different model families, we analyze the results across datasets as shown in Figure~\ref{fig:rank_dataset} and highlight some findings below.
% The discrimination between different model families is smaller than that between different micro-aggregation design dimensions by comparing Fig.~\ref{fig:dis_dataset} and Fig.~\ref{fig:gnn_type_dis}.

%We try to answer the question: \textbf{Are meta-paths or variants still useful in GNNs?}

\textbf{The meta-path model family helps node classification.}
In node classification task, the meta-path model family outperforms visibly than the other model families on datasets HGBn-ACM and HGBn-DBLP, where we think some informative and effective meta-paths have been empirically discovered. Some experimental analysis for meta-paths can be found in Appendix~\ref{app:meta-path}.

The meta-path model family does not help link prediction. In previous works, few variants of the meta-path model family were applied to the link prediction task. Although our unified framework can apply the meta-path model family to the link prediction task, the meta-path model family does not perform well on all datasets of link prediction as shown in Figure~\ref{fig:rank_dataset}. We think this is because the information from the edges in the original graph is important in link prediction, and the meta-path model family ignores it.

\textbf{The relation model family is a safer choice in all datasets.}
From Figure~\ref{fig:rank_dataset}, the relation model family stands out in link prediction task, which confirms the necessity to preserve the edges as well as their type information in the original graph. Compared with the homogenization model family, the relation model family has more trainable parameters which have a linear relationship with the number of edge types. Surprisingly, the relation model family is not 
very effective in HGBl-Freebase with much heterogeneity, which has 8 node types and 36 edge types. We think that too many parameters lead to over-fitting, which may challenge the relation model family.
According to the distribution of ranking, the relation model family has a significantly lower probability of being ranked last. Therefore, the relation model family is a safer choice.%From Fig.~\ref{fig:dis_dataset}, relation model family is neither outstanding nor backward and is a safer choice.

\textbf{The homogenization model family with the least trainable parameters is still competitive.}
As shown in Figure~\ref{fig:rank_dataset}, the homogenization model family is still competitive against the relation model family on HGBl-IMDB, and even outperforms the latter on HGBn-Freebase and HGBl-LastFM. Therefore, the homogenization model family is not negligible as a baseline even on heterogeneous graphs, which aligned with \cite{Simple-HGN}. %As a vanilla method in HGNN, it shows we still have space to improve HGNN.

\subsubsection{\textbf{The Micro-aggregation and the Macro-aggregation Design Dimensions}}
The existing HGNNs are usually inspired by GNNs and apply different micro-aggregation (e.g., GCNConv, GATConv). The micro-aggregation design dimension in our design space brings many variants to the existing HGNNs. As shown in Figure~\ref{fig:task_rank}, the results of comparison between micro-aggregation vary greatly across tasks. We provide ranking analysis on different datasets in Appendix~\ref{app:_micro}.
%An intuitive question is whether different micro-aggregation bring different effects. 
% As shown in Figure~\ref{fig:gnn_type_dis}, the results of comparison between micro-aggregation vary greatly across datasets. The GCNConv has gained significant advantages on datasets HGBl-amazon and HGBl-LastFM. The GATConv performs best on the two datasets. The SOTA model GINConv for graph-level tasks can also stand out in one dataset HGBn-DBLP here. It confirms that there is no single GNN model can perform well in all situations. 
%Therefore, it is of practical significance for us to distill the model families and derive the corresponding variants in the micro-aggregation design dimension.

For the macro-aggregation design dimension, Figure~\ref{fig:task_rank} shows that Sum has a great advantage in both tasks, which is aligned with the theory that \emph{Sum aggregation is theoretically most expressive} \cite{GIN}. Surprisingly, Attention is not so effective as Sum, and we think the micro-aggregation is powerful enough, resulting that complicated Attention in macro-aggregation is not necessary.

% The GCNConv has gained significant advantages in HGBl-amazon and HGBl-LastFM. The GATConv and SageConv-mean perform best on the two datasets respectively. The SOTA model GINConv for graph-level tasks can also stand out in one of the datasets HGBN-DBLP here. It seems that the simpler model achieves better performance than the more sophisticated in the same training procedure and search space, which is aligned with \cite{shchur2018pitfalls}. We think the simpler models are more robust in diverse situations. While this result seems surprising, similar findings have been reported in other fields \cite{melis2018on, GANs}. Therefore, it is of practical significance for us to distill the model families and derive the corresponding variants in the micro-aggregation design dimension.

% \begin{figure*}[htpb]
%     \centering
%     \includegraphics[width=\textwidth]{images/condensed_dis.png}
%     \caption{\textbf{Distribution estimates for different design space.}  Curves closer to the lower-right corner indicates a better design space.  The vertical dashed line indicates the best performance GraphNAS can get.}
%     \label{fig:condensed_dis}
% \end{figure*}

% In summary, the experimental results offer us guidelines to design and implement HGNN. We also answer a question from \cite{Simple-HGN} and verify that the meta-path model family and the relation model family are still useful. We encourage researchers to conduct more experiments and explore more findings with Space4HGNN.

\begin{table}[htpb]
    \centering
    \caption{Condensed common design dimensions with GraphGym.  Unique dimensions in HGNNs are not condensed.}
    \resizebox{\columnwidth}{!}{
    \begin{tabular}{c|c|c}
    \toprule
       \textbf{Design Dimension} & \textbf{\makecell[c]{Our Condensed \\Design Space}} & \textbf{\makecell[c]{Condensed Design Space\\ in GraphGym}} \\
    \midrule
        Batch Normalization & True, False & True\\
        Dropout & 0, 0.3 & 0 \\
        Activation & ELU LeakyReLU Tanh & PReLU \\
        L2 Normalization & True, False &  - \\
    \midrule
        Layer Connectivity & SKIP-SUM, SKIP-CAT & SKIP-SUM, SKIP-CAT \\
        Pre-process Layers & 1 & 1, 2\\
        Message Passing Layers & 1, 2, 3, 4, 5, 6 & 2, 4, 6, 8\\
        Post-process Layers & 1, 2 & 2, 3\\
    \midrule
        Optimizer & Adam & Adam\\
        Learning Rate & 0.1, 0.01 & 0.01\\
        Training Epochs & 400 & 400\\
        Hidden dimension & 64, 128 & -\\
    \bottomrule
    \end{tabular}}
    \label{tab:condensed}
\end{table}

\begin{table*}[htpb]
    \caption{Comparison with the standard HGNNs in HGB. The prefix HGBn means dataset of node classification and the prefix HGBl means dataset of link prediction. Vacant positions (-) are due to lack of baselines in HGB. The lower average rank is better.}
    \centering
    \resizebox{\textwidth}{!}{
    \begin{tabular}{c|cc|cc|cc|cc|cc|c}
    \toprule
         & \multicolumn{2}{c|}{HGBn-DBLP} & \multicolumn{2}{c|}{HGBn-ACM} &\multicolumn{2}{c|}{HGBl-amazon} &\multicolumn{2}{c|}{HGBl-LastFM}  \\
    \midrule
         & Macro-F1 & Micro-F1 & Macro-F1 & Micro-F1 & ROC-AUC & MRR & ROC-AUC & MRR  & Average Rank\\
    \midrule
        GCN & 90.84$\pm0.32$ & 91.47$\pm0.34$ & 92.17$\pm0.24$ & 92.12$\pm0.23$ & 92.84$\pm0.34$ & \textbf{97.05$\pm$0.12} & 59.17$\pm0.31$ & 79.38$\pm0.65$ & 3.75\\
    \midrule
        GAT & 93.83$\pm0.27$ & 93.39$\pm0.30$& 92.26$\pm0.94$ & 92.19$\pm0.93$ & 91.65$\pm0.80$ & 96.58$\pm0.26$ & 58.56$\pm0.66$ & 77.04$\pm2.11$ & 3.63\\
    \midrule
        RGCN & 91.52$\pm0.50$ & 92.07$\pm 0.50$ & 91.55$\pm 0.74$ & 91.41$\pm 0.75$ & 86.34$\pm0.28$& 93.92$\pm0.16$ & 57.21$\pm0.09$ & 77.68$\pm0.17$ & 5.13\\
    \midrule
        HAN & 91.67$\pm0.49$ & 92.05$\pm0.62$&  90.89$\pm 0.43$ & 90.79$\pm0.43$ & - & - & - & - & 6.63\\
    \midrule
        HGT & 93.01$\pm0.23$ &93.49$\pm0.25$ &91.12$\pm0.76$ &91.00$\pm0.76$& 88.26 $\pm2.06$ & 93.87 $\pm0.65$ & 54.99$\pm0.28$ & 74.96 $\pm1.46$ & 5.25\\
    \midrule
        Simple-HGN & 94.01$\pm0.24$ & 94.46$\pm 0.22$ & \textbf{93.42$\pm$0.44} & \textbf{ 93.35$\pm$0.45} & 93.49 $\pm0.62$ & 96.94$\pm0.29$ & 67.59$\pm0.23$ & 90.81$\pm0.32$ & 1.75\\
    \midrule
        Ours & \textbf{94.24$\pm$0.42} & \textbf{94.63$\pm$0.40}& 92.50 $\pm$0.14& 92.38$\pm$0.10& \textbf{95.15$\pm$0.43} & 96.56$\pm0.29$ & \textbf{70.15$\pm$0.77} & \textbf{91.12$\pm$0.38} & \textbf{1.63} \\
    \bottomrule
    \end{tabular}}
    \label{tab:NC_performance}
\end{table*}

\begin{table*}[htpb]
    \centering
    \caption{Best designs in HGB, found by a simple random search in our condensed design space. }
    \resizebox{\textwidth}{!}{
    \begin{tabular}{c|c|c|c|c|c|c|c|c|c|c}
    \toprule
        Dataset & Model Family & \makecell[c]{Micro-\\aggregation} & \makecell[c]{Macro-\\aggregation} & BN & L2-Norm & Dropout & Activation& \makecell[c]{Layer\\Connectivity} & \makecell[c]{Message\\Passing\\Layers} & \makecell[c]{Post-\\process \\Layers} \\
    \midrule
        HGBn-ACM & Relation & GCNConv & Sum & False & True & 0.0 & Tanh & SKIP-SUM & 2 & 2 \\
        HGBn-DBLP & Homogenization & GATConv & - & True & True & 0.0 & LeakyRelu & SKIP-SUM & 3 & 2\\
        HGBl-amazon & Meta-path & GATConv & Attention & True & False & 0.0 & LeakyRelu & SKIP-SUM & 5 & 1 \\
        HGBl-LastFM & Homogenization & SageConv & - & True & False & 0.3 & Elu & SKIP-CAT & 5 & 2 \\
    \bottomrule
    \end{tabular}}
    \label{tab:best_design}
\end{table*}
\subsection{Evaluation of Condensed Design Space}
Above experiments reveals that it is hard to design a single HGNN model that can guarantee outstanding performance across diverse scenarios in the real world. 
% Motivated by \cite{radosavovic2019network, radosavovic2020designing}, focusing on design space that consists of a collection of models instead of individual architecture brings more generalization and robustness. 
According to the findings in Section~\ref{exp:common_design}, we condensed the design space to facilitate model searching. 
Specifically, we remove some bad choices in design dimensions and retain some essential design dimensions (e.g., high-level architectural structures and helpful design principles).
The evaluation of HGB shows that a simple random search in the condensed design space can find the best designs. More experimental results compared GraphGym~\cite{you2020design} and GraphNAS~\cite{graphnas} are analyzed in Appendix~\ref{app:condensed}.

\subsubsection{The Condensed Design Space}
For common design dimensions with GraphGym, Table~\ref{tab:condensed} compares the condensed design spaces we and GraphGym proposed. We retain some of the design dimensions same as GraphGym if the findings in Section~\ref{exp:common_design} are aligned (i.e., layer connectivity, optimizer, and training epochs).  We propose our own choices for the different dimensions in conclusion (e.g., Dropout, activation, BN, L2-Norm, etc.).
%Following some findings specific to HGNN, we have a minor modification to some design dimensions (e.g., dropout, activation, and pre-process layers). Besides, there are some non-negligible results in task-wise evaluation, and we retained all choices for BN and L2-Norm.
For unique design dimensions in HGNN, we conclude that the micro-aggregation and model family design dimensions vary greatly across datasets or tasks. So we retain all choices in unique design dimensions and aim to find out whether the variants of existing HGNNs could gain improvements in HGB.

The original design space contains over 40M combinations, and the condensed design space contains 70K combinations. So the possible combination of the design dimensions in condensed design space is reduced by nearly 500 times.

\subsubsection{Evaluation in Heterogeneous Graph Benchmark (HGB)}
To compare with the performance of the standard HGNNs, we evaluate our condensed design space in a new benchmark HGB. We randomly searched 100 designs from condensed design space and evaluated the best design of validation set in HGB. As shown in Table~\ref{tab:NC_performance}, our designs with condensed design space can achieve comparable performance. So we can easily achieve SOTA performance with a simple random search in the condensed design space. Table~\ref{tab:best_design} shows the best designs we found in our condensed design space, which cover the variants of RGCN and HAN. It also confirms that the meta-path model family and the relation model family have great performance in HGB and answers the question \emph{``are meta-path or variants still useful in GNNs?"} from \cite{Simple-HGN}.
Note that this result does not contradict the conclusion from \cite{Simple-HGN}, as our design space includes much more components than the vanilla RGCN or HAN model, and proper components can make up shortcomings of an existing model.

\section{Conclusion and Discussion}
In this work, we propose a unified framework of HGNN and define a design space for HGNN, which offers us a module-level sight to evaluate HGNN models. Specifically, we comprehensively analyze the common design dimensions with GraphGym and the unique design dimensions in HGNN. After that, we distill some findings and condense the original design space. Finally, experimental results show that our condensed design space outperforms others, and gains the best average ranking in a benchmark HGB. With that, we demonstrate that focusing on the design space could help drive advances in HGNN research.

\textbf{How to condense the design space?}
In our work, the condensed design space is distilled according to the findings within extensive experiments, which still needs much effort and intuitive experience. A recently proposed work KGTuner~\cite{zhang2022kgtuner}, which analyzed the design space for knowledge graph embedding, proposed a more systematic way to shrink and decouple the search space, which can be a potential improvement of this work
\begin{acks}
This work is supported in part by the National Natural Science Foundation of China (No. U20B2045, 62192784, 62002029, 62172052, 61772082).
\end{acks}

% %%
% %% The next two lines define the bibliography style to be used, and
% %% the bibliography file.

% \appendix
% \section*{Appendix}
% For sake of the space, the appendix including preliminary, design space and datasets description, more technical details, and experimental results is provided in: \url{https://anonymous.4open.science/r/Space4HGNN-862F}.

\bibliographystyle{ACM-Reference-Format}
\bibliography{sample-sigplan}

% %%
% %% If your work has an appendix, this is the place to put it.
\appendix

\section{Preliminary}
\label{app:preliminary}
\subsection{Graph Neural Network}
Graph Neural Networks (GNNs) aim to apply deep neural networks to graph-structured data. Here we focus on message passing GNNs which could be implemented efficiently and proven great performance. 
\begin{definition}[Message Passing GNNs \cite{messagepssing}]
Message passing GNNs aim to learn a representation vector $\mathbf{h}_{v}^{(L)} \in \mathbb{R}^{d_{L}}$ for each node $v$ after $L$-th message passing layers of transformation, and $d_{L}$ means the output dimension in $L$-th message passing layer.
% Let $\boldsymbol{h}_{v} \in \mathbb{R}^{d}$ be the feature for node $v$. 
The message passing paradigm defines the following node-wise and edge-wise computation for each layer as:
\begin{equation}
\text { Edge-wise: } \mathbf{m}_{e_{i j}}^{(L+1)}=\phi\left(\mathbf{h}_{i}^{(L)}, \mathbf{h}_{j}^{(L)}\right), j \in \mathcal{N}_{i},
\end{equation}
% \quan{What is $w_e^{(L)}$?  You only defined $w_e$.}
where $\mathcal{N}_{i}$ means neighbors of node $v_i$, $\phi$ is a message function defined on each edge to generate a message by combining the features of its incident nodes, and $e_{i j}$ denotes an edge from node $v_j$ to $v_i$;

\begin{equation}
\text { Node-wise: } \mathbf{h}_{i}^{(L+1)}=\psi\left(\mathbf{h}_{i}^{(L)}, \rho\left(\left\{\mathbf{m}_{e_{i j}}^{(L+1)}: \forall{j} \in \mathcal{N}_{i}\right\}\right)\right),
\end{equation}
where $\mathcal{N}_{i}$ means neighbors of node $v_i$, $\psi$ is an update function defined on each node to update the node representation by aggregating its incoming messages using the aggregation function $\rho$.  
% \quan{The message set is wrong; it should only contain the neighbors, not the entire $\mathcal{E}$.}
\end{definition}
\textbf{Example:} GraphSAGE~\cite{sage} can be formalized as a message passing GNN, where the message function is $\phi = h_{j}^{(L)} W^{(L)}$ and the update function is $\psi = \operatorname{SUM}{\left(\{ \mathbf{m}_{e_{i j}}^{(L+1)}: \forall{j} \in \mathcal{N}_{i}\}\right)}$. 
% \quan{[Agree]Same problem.  Also for GCN you need to multiply messages with $1/\sqrt{d_i d_j}$.  Or you could replace the example with GraphSAGE.}

\subsection{Heterogeneous Graph}
\begin{definition}[Heterogeneous Graph]
A heterogeneous graph, denoted as  $\mathcal{G}=(\mathcal{V},\mathcal{E})$ , consists of a node set $\mathcal{V}$ and an edge set $\mathcal{E}$. A heterogeneous graph is also associated with a node type mapping function $f_{v}: \mathcal{V} \rightarrow \mathcal{T}^{v}$ and an edge type (or relation type) mapping function $f_{e}: \mathcal{E} \rightarrow \mathcal{T}^{e}$. $\mathcal{T}^{v}$ and $\mathcal{T}^{e}$ denote the sets of node types and edge types.
Each node $v_{i} \in \mathcal{V}$ has one node type $f_{v}\left(v_{i}\right) \in \mathcal{T}^{v}$. Similarly, for an edge $e_{i j} \in \mathcal{E}$ from node $i$ to node $j$, $f_e(e_{i j}) \in \mathcal{T}^e$.
% for an edge  $e_{i j} \in \mathcal{E}$ from node $i$ to node$j$, $f_{e}\left(e_{i j}\right) \in \mathcal{T}^{e}$. 
When $\left|\mathcal{T}^{v}\right|>1$ or $\left|\mathcal{T}^{e}\right|>1$, it is a heterogeneous graph, otherwise it is a homogeneous graph.  
% \quan{A few suggestions: (1) $\mathbf{X}$ is not used in the definition, while $\mathcal{T}^v, \mathcal{T}^e, f_v, f_e$ did not appear in the text before.  (2) The use of $\mathcal{E}$ is inconsistent: here it is a set of edges while in the above section it is a set of source-destination-edge triplets.  I would change this sentence to "for an edge $(i, j, e) \in \mathcal{E}$ from node $i$ to node $j$, $f_e(e) \in \mathcal{T}^e$.  (3) For prettiness and reference I would suggest you use \texttt{amsthm} package for writing definitions/theorems/etc.  See \url{https://www.overleaf.com/learn/latex/Theorems_and_proofs}.}
\end{definition}
\textbf{Example.} As shown in Figure~\ref{fig:architecture} (left), we construct a simple heterogeneous graph to show an academic network. It consists of multiple types of objects (Paper(P), Author (A), Conference(C)) and relations (written-relation between papers and authors, published-relation between papers and conferences).
\begin{definition}[Relation Subgraph] \label{relation subgraph}
A heterogeneous graph can also be represented by a set of adjacency matrices $\left\{A_{k}\right\}_{k=1}^{K}$, where $K$ is the number of edge types $\left|\mathcal{T}^{e}\right|$. $A_{k} \in \mathbf{R}^{n_s \times n_t}$ is an adjacency matrix where $A_{k}[i, j]$ is non-zero when there is an edge $e_{i j}$ with $k$-th type from node $v_{i}$ to node $v_{j}$. $n_{s}$ and $n_{t}$ are numbers of source and target nodes corresponding to the edge type $k$ respectively. A relation subgraph of $k$-th edge type is therefore a subgraph whose adjacency matrix is $A_k$.
% \quan{[Agree]A bit confusing here: an adjacency matrix is a subgraph?  You can say "A relation subgraph of edge type $k$ is therefore a subgraph whose adjacency matrix is $A_k$" (if I got it right).} 
As shown in Figure~\ref{fig:graph} (b), the underlying data structure of the academic network in Figure~\ref{fig:architecture} (left) consists of four adjacency matrices.
\end{definition}

\begin{definition}[Meta-path \cite{sun2011pathsim}]\label{MP}
A meta-path $\mathcal{P}$ is defined as a path in the form of $v_{1} \stackrel{r_{1}}{\longrightarrow} v_{2} \stackrel{r_{2}}{\longrightarrow} \cdots \stackrel{r_{l}}{\longrightarrow} v_{l+1}$ which describes a composite relation $r_{1} \circ r_{2} \circ \cdots \circ r_{l}$ between two nodes $v_{1}$ and $v_{l+1}$, where $r_{l} \in \mathcal{T}^{e}$ denotes the $l$-th relation type of meta-path and $\circ$ denotes the composition operator on relations.

\begin{definition}[Meta-path Subgraph] \label{mp_subg}
Given a meta-path $\mathcal{P}$, $r_{1} \circ r_{2} \circ \cdots \circ r_{l}$ , the adjacency matrix $A_{\mathcal{P}}$ can be obtained by 
a multiplication of adjacency matrices according relations as
\begin{equation}
\boldsymbol{A}_{\mathcal{P}}=\boldsymbol{A}_{r_{1}} \ldots \boldsymbol{A}_{r_{l-1}} \boldsymbol{A}_{r_{l}}.
\label{eq1}
\end{equation}
% \quan{It's better to split this into two definitions: meta-path (which you need to cite the original meta-path paper) and meta-path subgraph (also define it).}
\end{definition}
The notion of meta-path subsumes multi-hop connections and a \textbf{meta-path subgraph} is multiple relation subgraphs matrices multiplication shown in Figure~\ref{fig:graph} (b) (iii). So relation subgraph is a special case of meta-path subgraph which is only composited by a relation subgraph. When meta-path beginning node type and ending node type are the same, meta-path subgraph is a homogeneous graph, otherwise a bipartite graph. 
% \quan{A meta-path subgraph is not really a composition of multiple subgraphs, so this paragraph is inaccurate.  I will just remove it though, and simply say "Note that relation subgraph is a special case of meta-path subgraph" and stop.}
\end{definition}

\section{Design Space}
\label{app:design}
\subsection{Common Design with GraphGym}

\paragraph{\textbf{Intra-layer}}
Same with GNNs, an HGNN contains several Heterogeneous GNN layers, where each layer could have diverse design dimensions. As illustrated in Figure~\ref{fig:design}, the adopted Heterogeneous GNN layer has an aggregation layer which involves unique design dimensions discussed later, followed by a sequence of modules: 
(1) batch normalization BN($\cdot$) \cite{BN}; 
(2) dropout DROP($\cdot$) \cite{dropout}; 
(3) nonlinear activation function ACT($\cdot$); 
(4) L2 Normalization L2-Norm($\cdot$). 
Formally, the L-th heterogeneous GNN layer can be defined as:
% \begin{equation}
% \begin{aligned}
% \mathbf{h}_{v}^{(L+1)}=\operatorname{Inter-layer}( \operatorname{AGG}\left\{\mathbf{h}_{u}^{(L)}, u \in \mathcal{N}_{v})\right\}),
% \end{aligned}
% \end{equation}

% \begin{equation}
%     \operatorname{Inter-layer(\cdot)} = \operatorname{L2-Norm}\left(\operatorname{ACT}\left(\operatorname{DROP}\left(\operatorname{BN}\left( \cdot \right)\right)\right)\right).
% \end{equation}
\begin{small}
\begin{equation}
\begin{aligned}
\mathbf{h}_{v}^{(L+1)}=\operatorname{L2-Norm}\left(\operatorname{ACT}\left(\operatorname{DROP}\left(\operatorname{BN}\left(
\operatorname{AGG}\left\{\mathbf{h}_{u}^{(L)}, u \in \mathcal{N}_{v})\right\}
\right)\right)\right)\right).
\end{aligned}
\end{equation}
\end{small}

\paragraph{\textbf{Inter-layer}}
The layers of message passing, pre-processor and post-processor are supposed to be considered, which are essential design dimensions according to empirical evidence from neural networks. HGNNs face the problems of vanishing gradient, over-fitting and over-smoothing, and the last problem is seen as the obstacles to stack deeper GNN layers. Inspired by ResNet \cite{resnet} to alleviate the problems, skip connection \cite{deepgcns, xu2018representation} has been proven to significant effect. Therefore, we investigate two choices of skip connections: SKIP-SUM \cite{resnet} and SKIP-CAT \cite{huang2017densely} with STACK as a basic comparison.
    
\paragraph{\textbf{Training Settings}}
As a part of deep learning, we also want to analyze design dimensions on training settings, like optimizer, learning rate and training epochs. Besides, the hidden dimension is also included here involving the trainable parameters.

\begin{table*}[htpb]
    \caption{\textbf{Statistics of HGB datasets.} The prefix \emph{HGBn} presents datasets in the node classification task and target node with number of classes is for these datasets. The prefix \emph{HGBl} presents datasets in the link prediction task and target link is for these datasets.}
    \label{app_tab:dataset Statistics}
    \centering
    \begin{tabular}{c|c|c|c|c|c|c|c|c|c}
    \toprule
        Dataset & \#Nodes & \makecell[c]{\#Node \\Types} & \#Edges & \makecell[c]{\#Edge\\ Types} & \makecell[c]{Name for Node \\ Classification Task}& Target Node & \#Classes  & \makecell[c]{Name for Link\\ Prediction Task}&\makecell[c]{Target Link for \\ Link Prediction}  \\
    \midrule
        DBLP & 26,128 & 4 & 239,566 & 6 & HGBn-DBLP & author & 4 & HGBl-DBLP &author-paper \\ 
        IMDB & 21,420 & 4 & 86,642 & 6 & HGBn-IMDB & movie & 5 & HGBl-IMDB &actor-movie \\
        ACM & 10,942 & 4 & 547,872 & 8 & HGBn-ACM & paper & 3 & HGBl-ACM &paper-paper\\
        Freebase & 180,098 & 8 & 1,057,688 & 36 & HGBn-Freebase & book & 7 &-& -\\
        PubMed & 63,109 & 4 & 244,986 & 10 & HGBn-PubMed & disease & 8 & HGBl-PubMed &disease-disease \\
        Amazon & 10,099 & 1 & 148,659 & 2 &-&-&-&HGBl-amazon&product-product\\
        LastFM & 20,612 & 3 & 141,521 & 3 &-&-&-&HGBl-LastFM &user-artist \\
    % \midrule
    %     \makecell[c]{Link\\Prediction} & \multicolumn{4}{c|}{} & \multicolumn{2}{c}{target link} \\
    % \midrule
    %     Amazon & 10,099 & 1 & 148,659 & 2 & \multicolumn{2}{c}{product-product} \\
    %     LastFM & 20,612 & 3 & 141,521 & 3 & \multicolumn{2}{c}{user-artist} \\
    %     PubMed & 63,109 & 4 & 244,986 & 10 & \multicolumn{2}{c}{disease-disease} \\
    \bottomrule
    \end{tabular}
\end{table*}
\section{Dataset}
We select datasets from the Heterogeneous Graph Benchmark (HGB) \cite{Simple-HGN}, a benchmark with multiple datasets of various heterogeneity (i.e., the number of nodes and edge types), for node classification and link prediction tasks. 
The HGB is organized as a public competition, so it does not release the test label to prevent data leakage. Since formal submission to the public leaderboard costs a large amount of time and submission resources, we only report the test performance of the configuration with the best validation performance in Table~\ref{tab:NC_performance}, using the same metrics as in \cite{Simple-HGN}. Other experiments are evaluated on a validation set with three random 80-20 training-validation splits. 
The statistics of HGB are shown in Table~\ref{app_tab:dataset Statistics}. We select five datasets (DBLP, IMDB, ACM, Freebase, PubMed) for node classification task, six datasets (DBLP, IMDB, ACM, amazon, LastFM, PubMed) for link prediction task.
% \quan{Did I get this correctly?}
%If we get performance by submitting results, our experiments will cost a amount of time and submission resources even though the HGB Team gives us a test channel to implement the evaluation of test data. So we only evaluate the best performance in the test data in Table ~\ref{tab:NC_performance}, and other experiments will spilt the original train data into train/valid 80\%/ 20\%  and repeat 3 times randomly. For dataset details, please refer to Appendix ~\ref{app:dataset}. 

\section{Evaluation of Unique Design Dimensions}
\subsection{Analysis for Meta-path}
\label{app:meta-path}
\begin{table}[htpb]
    \centering
    \caption{Homogeneous subgraph extracted by meta-paths or relations and the corresponding homophily $\beta$ (bold is highest in the dataset).}
    \begin{tabular}{l|l|c|c}
        \toprule
        \textbf{Dataset} & \textbf{Meaning} & \textbf{Meta-path} & \textbf{$\beta$}  \\
        \midrule
        \multirow{8}{*}{\textbf{HGBn-ACM}} & \multirow{8}{*}{\makecell[l]{P: paper \\ A: author \\  S: subject \\ c: citation relation \\ r: reference relation}} &  PrP & 0.4991 \\
        && PcP & 0.4927 \\
        && PAP & \textbf{0.6511} \\
        && PSP & 0.4572 \\
        && PcPAP & 0.5012 \\
        && PcPSP & 0.4305 \\
        && PrPAP & 0.4841 \\
        && PrPSP & 0.4204 \\
        \midrule
        
        \multirow{3}{*}{\textbf{HGBn-DBLP}} & \multirow{3}{*}{\makecell[l]{A: author  P: paper \\ T: term  V: venue}} & APA & \textbf{0.7564} \\
        && APTPA & 0.2876 \\
        && APVPA & 0.3896 \\
        \midrule
        
        \multirow{5}{*}{\textbf{HGBn-PubMed}} &\multirow{5}{*}{\makecell[l]{D: disease \\ G: gene \\ C: chemical \\ S: species}} & DD &0.0169 \\
        && DCD &  0.1997 \\
        && DDD & 0.1945 \\
        && DGD & \textbf{0.2567} \\
        && DSD & 0.2477 \\
        \midrule
        \multirow{7}{*}{\textbf{HGBn-Freebase}} &\multirow{7}{*}{\makecell[l]{B: book F: film \\ L: location \\ M: music \\ P: person \\ S: sport \\ O: organization \\ U: business}} & BB & 0.1733 \\
        && BUB &  0.0889 \\
        && BFB & 0.1033 \\
        && BLMB & 0.0303 \\
        && BOFB & \textbf{0.3341} \\
        && BPB & 0.1928 \\
        && BPSB & 0.0603\\
        \bottomrule
    \end{tabular}
    \label{tab:homophily}
\end{table}

In node classification task, the meta-path model family outperforms visibly than the other model families on datasets HGBn-ACM and HGBn-DBLP, where we think some informative and effective meta-paths have been empirically discovered.
The micro-aggregation modules are MPNN networks that tend to learn similar representations for proximal nodes in a graph \cite{pei2019geom}. Moreover, the meta-path model family aims to bring nodes with the same type topologically closer with meta-path subgraph extraction, hoping that the extracted subgraph is assortative (e.g., citation networks) where node homophily holds (i.e., nodes with the same label tend to be proximal, and vice versa).
%The message passing mechanism is inclined to learn similar representations for proximal nodes in a graph \cite{pei2019geom}. The meta-path model family aims to explicitly shorten the distance between such nodes that are more likely to be similar. For example, the distance between two nodes with co-author relation will be shortened from two hops to one hop with meta-path subgraph extraction.
% The hypothesis that makes the meta-path model family work is that the subgraphs extracted by the meta-path are assortative graphs (e.g. citation networks) where node homophily holds (i.e., similar nodes are more likely to be proximal, and vice versa).
%An intuitive idea is that the reason for the poor performance of the meta-path model family is that the extracted sub-graphs are disassortative graphs \cite{newman2002assortative} where node homophily does not hold. 
Based on that, we measure the homophily \cite{pei2019geom} in subgraphs extracted by meta-path $\mathcal{P}$, which is defined as
% \begin{equation}
% \beta=\frac{1}{|V|} \sum_{v \in V} \frac{ \makecell[c]{ \# v \text { 's neighbors} \\ \text {with the same label as } v}}{ \# v \text {'s neighbors}}.
% \end{equation}
\begin{equation}
\beta = \frac{1}{|V|} \sum_{v \in V}
\frac{| \{ u: \boldsymbol{A}_{\mathcal{P}}[u, v] = 1, y_u = y_v \}|}{| \{ u: \boldsymbol{A}_{\mathcal{P}}[u, v] = 1\}|},
\end{equation}
where $y_u$ and $y_v$ represent the label of node $u$ and $v$, respectively.

%A large $\beta$ value implies strong homophily
%that the homophily, in term of node label, is strong in a graph, i.e., similar nodes tend to connect together.
As shown in Table~\ref{tab:homophily}, the homophily of homogeneous subgraphs extracted by predefined meta-path in HGBn-ACM and HGBn-DBLP is significantly higher than that in HGBn-PubMed and HGBn-Freebase. For node classification task, the homophily of subgraphs extracted by meta-paths may be a helpful reference for meta-path selection. So for the question \emph{``are meta-path or variants still useful in GNNs?"} from \cite{Simple-HGN}, we think that the meta-path model family is still useful with well-defined meta-paths that reveal task-specific semantics.% with good meta-paths, which could help models discover the effective information for the specific task.

\subsection{Analysis for Micro-aggregation}
\label{app:_micro}
\begin{figure*}[htpb]
    \centering
    \includegraphics[width=\textwidth]{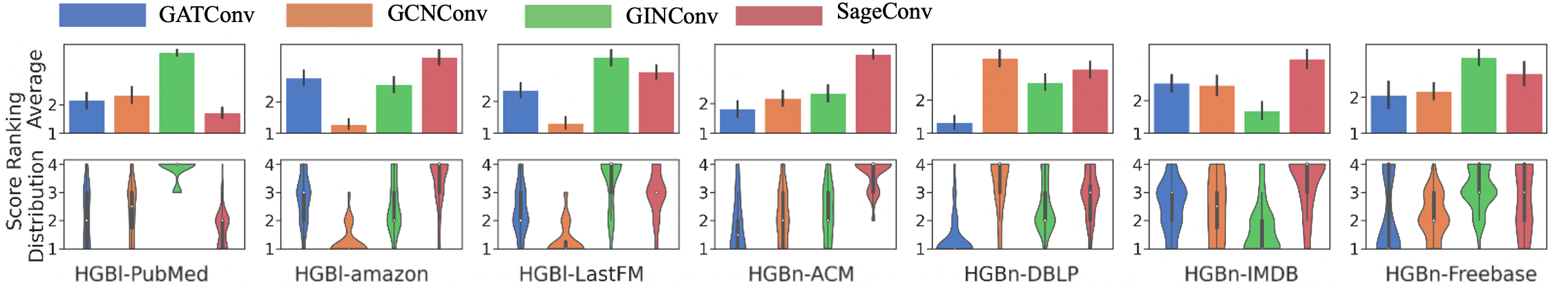}
    \caption{\textbf{Ranking analysis for the micro-aggregation design dimension on different datasets.} Different micro-aggregation vary greatly across datsest.}
    \label{fig:gnn_type_dis}
\end{figure*}
As shown in Figure~\ref{fig:gnn_type_dis}, the results of comparison between micro-aggregation vary greatly across datasets. The GCNConv has gained significant advantages on datasets HGBl-amazon and HGBl-LastFM. The GATConv performs best on the two datasets. The SOTA model GINConv for graph-level tasks can also stand out in one dataset HGBn-DBLP here. It confirms that there is no single GNN model can perform well in all situations. 

\begin{figure*}[htpb]
    \centering
    \includegraphics[width=\textwidth]{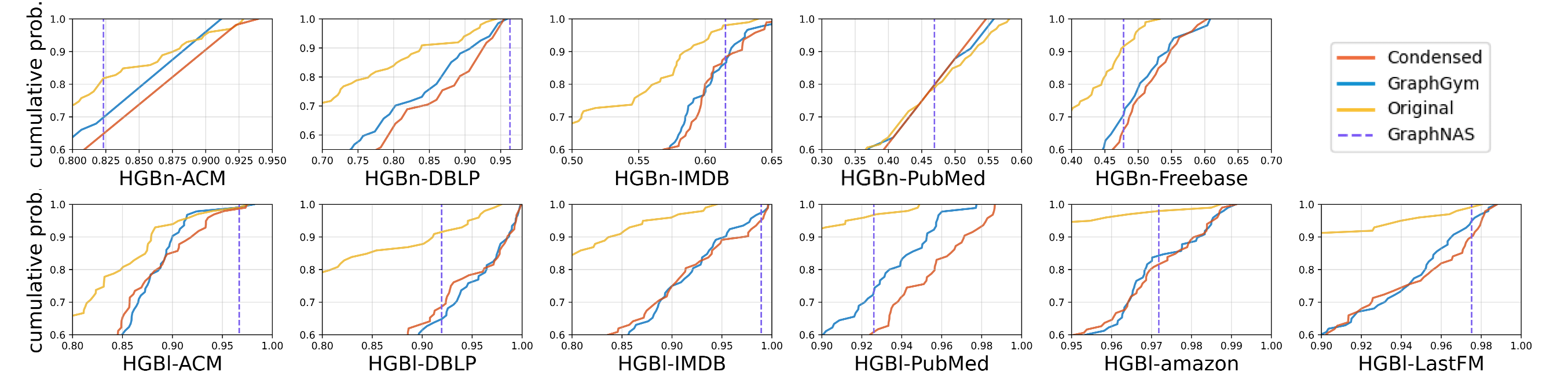}
    \caption{\textbf{Distribution estimates for different design space.}  Curves closer to the lower-right corner indicates a better design space.  The vertical dashed line indicates the best performance GraphNAS can get.}
    \label{fig:condensed_dis}
\end{figure*}

\section{Evaluation of condensed design space}
\label{app:condensed}
\subsection{Evaluation of Different Design Spaces}
Plotting ranking with controlled random search can only work in the same design space, and is not suitable for evaluation across different design spaces.  Therefore, we plot for each design space the empirical distribution function (EDF) \cite{radosavovic2019network}: given $n$ configurations and their respective scores $s_i$, EDF is defined as
\begin{equation}
F(s)=\frac{1}{n} \sum_{i=1}^{n} \mathbf{1}\left[s_{i}<s\right].
\end{equation}
EDF essentially tells the probability of a random hyperparameter configuration that \emph{cannot} achieve a given performance metric.  Therefore, with $x$-axis being the performance metric and $y$-axis being the probability, an EDF curve closer to the lower-right corner indicates that a random configuration is more likely to get a better result.

Though the condensed design space in GraphGym is small enough to perform a full grid search for GNNs, it is not so much suitable for HGNNs due to more complicated HGNN models. To verify the effectiveness of our condensed design space, we compare it with the original design space and the condensed design from GraphGym \cite{you2020design}. We randomly search 100 designs in three spaces, respectively.

As shown in Figure~\ref{fig:condensed_dis}, our condensed design space outperforms the others. Specifically, the original design space has many bad choices (i.e., optimizer with SGD) and performs worst in the distribution estimates. On the other hand, the best design in the original design space is competitive, but at a much higher search cost.  %not so bad, which means random search in the original design space could get a not so bad result but cost too much.
Besides, the better performance in our condensed design space compared with GraphGym shows that we cannot simply transfer the design space condensed from homogeneous graphs to HGNNs, and specific condensation is required.%it is imperative to define design space for HGNN and distill specific findings in HGNNs. 

% \begin{table}[h]
%     \centering
%     \caption{Condensed common design dimensions with GraphGym.  Unique dimensions in HGNNs are not condensed.}
%     \resizebox{\columnwidth}{!}{
%     \begin{tabular}{c|c|c}
%     \toprule
%       \textbf{Design Dimension} & \textbf{\makecell[c]{The original \\Design Space}} & \textbf{\makecell[c]{The Condensed \\Design Space}} \\
%     \midrule
%         Batch Normalization & True, False & True, False\\
%         Dropout & 0, 0.3, 0.6 & 0, 0.3 \\
%         Activation & Relu, LeakyRelu, Elu, Tanh, PRelu & ELU LeakyReLU Tanh  \\
%         L2 Normalization & True, False & True, False  \\
%     \midrule
%         Layer Connectivity & STACK,SKIP-SUM, SKIP-CAT & SKIP-SUM, SKIP-CAT \\
%         Pre-process Layers & 1, 2, 3 & 1\\
%         Message Passing Layers & 1, 2, 3, 4, 5, 6 & 1, 2, 3, 4, 5, 6\\
%         Post-process Layers & 1, 2, 3 & 1, 2\\
%     \midrule
%         Optimizer & Adam, SGD & Adam\\
%         Learning Rate & 0.1, 0.01, 0.001, 0.0001 & 0.1, 0.01\\
%         Training Epochs & 100, 200, 400 & 400\\
%         Hidden dimension & 8, 16, 32, 64, 128 & 64, 128\\
%     \bottomrule
%     \end{tabular}}
%     \label{ta}
% \end{table}

\subsection{Comparison with GraphNAS}
We also apply a GNN neural architecture search method GraphNAS \cite{graphnas} in the original design space as a comparison. The NAS is to find the best architecture, so we only report the best performance of GraphNAS in Figure~\ref{fig:condensed_dis}. Though GraphNAS outperforms in HGBn-DBLP and gains excellent performance in HGBl-ACM and HGBl-IMDB, it performs worst in other datasets. So compared with GraphNAS, our design space has more significant  advantages in robustness and stability. We think that we need a more advanced NAS method (i.e., DiffMG \cite{diffmg}) for our design space in future work.

% \appendix









\end{document}